\documentclass[10pt, a4paper,fleqn]{cas-dc}

\usepackage[numbers]{natbib}
\usepackage{soul}
\usepackage[normalem]{ulem} 
\usepackage{xcolor}         
\usepackage{amsmath}
\usepackage{amssymb}
\usepackage{placeins}  
\usepackage{graphicx}
\usepackage{hyperref}
\usepackage{caption}
\usepackage{subcaption}
\usepackage{tabularx}
\usepackage{todonotes}
\usepackage{svg}
\usepackage{enumitem}

\definecolor{mygreen}{RGB}{44, 160, 44}

\newcolumntype{Y}{>{\centering\arraybackslash}X}

\def\tsc#1{\csdef{#1}{\textsc{\lowercase{#1}}\xspace}}
\tsc{WGM}
\tsc{QE}
\tsc{EP}
\tsc{PMS}
\tsc{BEC}
\tsc{DE}

\begin{document}
\let\WriteBookmarks\relax
\def\floatpagepagefraction{1}
\def\textpagefraction{.001}



\title [mode = title]{Enhancing Glucose Level Prediction of ICU Patients through Hierarchical Modeling of Irregular Time-Series}




\author[1]{Hadi Mehdizavareh}

\cormark[1]


\ead{mhme@cs.aau.dk}


\affiliation[1]{organization={Department of Computer Science, Aalborg University},
    city={Aalborg},
    country={Denmark}}

\author[1]{Arijit Khan}
\ead{arijitk@cs.aau.dk}

\author[2]{Simon Lebech Cichosz}
\ead{simcich@hst.aau.dk}


\affiliation[2]{organization={Department of Health Science and Technology, Aalborg University},
    city={Aalborg},
    country={Denmark}}

\cortext[cor1]{Corresponding author}


\begin{abstract}
Accurately predicting blood glucose (BG) levels of ICU patients is critical, as both hypoglycemia (BG < 70 mg/dL) and hyperglycemia (BG > 180 mg/dL) are associated with increased morbidity and mortality. This study presents a proof-of-concept machine learning framework, the Multi-source Irregular Time-Series Transformer (\textsf{MITST}), designed to predict BG levels in ICU patients. In contrast to existing methods that rely heavily on manual feature engineering or utilize limited Electronic Health Record (EHR) data sources, \textsf{MITST} integrates diverse clinical data—including laboratory results, medications, and vital signs—without predefined aggregation. The model leverages a hierarchical Transformer architecture, designed to capture interactions among features within individual timestamps, temporal dependencies across different timestamps, and semantic relationships across multiple data sources. Evaluated using the extensive eICU database (200,859 ICU stays across 208 hospitals), \textsf{MITST} achieves a statistically significant (\( p < 0.001 \)) average improvement of 1.7 percentage points (pp) in AUROC and 1.8 pp in AUPRC over a state-of-the-art random forest baseline. Crucially, for hypoglycemia—a rare but life-threatening condition—\textsf{MITST} increases sensitivity by 7.2 pp, potentially enabling hundreds of earlier interventions across ICU populations. The flexible architecture of \textsf{MITST} allows seamless integration of new data sources without retraining the entire model, enhancing its adaptability for clinical decision support. While this study focuses on predicting BG levels, we also demonstrate \textsf{MITST}'s ability to generalize to a distinct clinical task (in-hospital mortality prediction), highlighting its potential for broader applicability in ICU settings. \textsf{MITST} thus offers a robust and extensible solution for analyzing complex, multi-source, irregular time-series data.
\end{abstract}


\maketitle

\section{Introduction}

The Intensive Care Unit (ICU) represents a critical care environment where timely and accurate decision-making is essential for patients' survival \cite{lighthall_understanding_2015}. Patients admitted to the ICU often suffer from complex, multifaceted conditions, requiring constant monitoring and rapid intervention \cite{jaderling_icu_2013, mitchell_prospective_2010, moss_signatures_2016}. ICU patients are typically subjected to continuous monitoring of various physiological parameters, including heart rate, blood pressure, oxygen saturation, and other critical indicators such as blood glucose (BG) levels \cite{saeed_mimic_2002}. The volume, velocity, and variety of data generated in the ICU can be overwhelming for clinicians to process in real-time, leading to potential delays in recognizing deteriorating conditions or impending adverse events \cite{portet_automatic_2009, hyland_early_2020}.

Hypoglycemia, characterized by BG levels below 70 mg/dL, and hyperglycemia, defined as BG levels exceeding 180 mg/dL, are both critically important in clinical management due to their association with increased morbidity and mortality \cite{zale_machine_2022, mujahid_machine_2021, diouri_hypoglycaemia_2021, cichosz_spontaneous_2019, holt_management_2021}. Patients with diabetes are particularly vulnerable to these glucose fluctuations because of impaired glycemic regulation. The global prevalence of type 2 diabetes mellitus (T2DM) is alarmingly high, affecting approximately 462 million individuals worldwide and imposing substantial burdens on healthcare systems \cite{khan_epidemiology_2020}.

Research indicates that in hospital settings, approximately 38-40\% of patients with T2DM experience hyperglycemic episodes, while 12-30\% face episodes of hypoglycemia \cite{dhatariya_management_2000, miller_hypoglycemia_2001}. Both conditions are strongly associated with prolonged hospital stays and increased mortality rates. Hypoglycemic events, particularly those occurring during night-time, are of even greater concern due to their significantly higher risk of fatal outcomes \cite{berikov_machine_2022}. Another study demonstrated that hyperglycemia contributes to 2.5\% increase in mortality and extends the median ICU stay by 0.4 days among diabetic ICU patients \cite{cichosz_hyperglycemia_2017}.

Conventional methods of glucose management often rely on periodic measurements and reactive treatments, which may not be optimal for maintaining glycemic control in a dynamic and high-risk setting like the ICU. A significant challenge of current literature on BG prediction task is their limitations in managing the highly irregular nature of electronic health record (EHR) data, as these methods typically require regular time interval-based measurements for accurate forecasting. Also, contemporary learning algorithms applied to this task often rely on aggregating the irregular EHR data to fit the model's needs \cite{zale_development_2022}. Unfortunately, predefined aggregation schemes tend to overlook crucial temporal information, which can contain significant insights into a patient's condition. For instance, aggregating data over fixed intervals can obscure critical patterns and trends that occur over shorter or varying timescales. 

Additionally, EHR data combine information from multiple sources, including medications, lab results, diagnoses, and patient-generated data. Each source has its own data format and characteristics, which presents several challenges:
\begin{itemize}
    \item \textbf{Integration}: Combining data having raw formats and arriving from different sources is complex due to differences in data structures, measurement units, and temporal resolution. This leads to incomplete or inconsistent datasets that are difficult to analyze.
    \item \textbf{Feature engineering}: Extracting meaningful features from raw data across multiple sources is challenging. Transforming raw features into a unified representation that captures the necessary information without losing important details is also difficult.
    \item \textbf{Lack of a shared representation}: EHR data encompass a wide range of concepts and entities, such as diagnoses, medications, procedures, and patient demographics. The lack of a standardized representation of these diverse elements makes it difficult to achieve semantic interoperability.  Developing unified models that can simultaneously handle diverse medical entities and their relationships is difficult. Without a shared representation, it is challenging to create models that understand and leverage the complex interdependencies within the medical data. Also, incorporating external medical knowledge (e.g., clinical guidelines, biomedical ontologies) into EHR data analysis is complicated by the lack of standardized representations. This limits the ability to enrich EHR data with additional contextual information that could enhance analysis and decision-making.
    \item \textbf{Computational complexity}: Analyzing massive data from multiple sources requires substantial computational resources and sophisticated algorithms to manage high dimensionality and heterogeneity.
\end{itemize}

Overcoming these challenges could enhance the accuracy, interoperability, and clinical relevance of EHR data analysis, ultimately improving healthcare outcomes. 
To this end, we propose \textsf{MITST} (an abbreviation for \textbf{M}ulti-source \textbf{I}rregular \textbf{T}ime-\textbf{S}eries \textbf{T}ransformer), a novel method designed to address the irregular nature of EHR data, while preserving essential temporal details often lost with conventional approaches. \textsf{MITST} integrates multiple sources of medical information, creating shared representations of medical concepts and entities. Our framework provides a more robust and insightful analysis of irregular time-series forecasting with a focus on timely BG level prediction tasks, improving patient outcomes in critical care settings. Our contributions are summarized as follows.
\begin{itemize}
    \item We propose an attention-based neural network solution called \textsf{MITST}, designed to predict BG levels for ICU patients using irregular clinical time-series data (\hyperref[fig:method]{Fig.~\ref*{fig:method}}). \textsf{MITST} addresses the challenge of predefined aggregation by eliminating the need for it and instead employs learning-based aggregation at different levels, capturing temporal details lost in conventional approaches.
    \item Our framework integrates multiple data sources (e.g., medication, lab results, vital signs) and handles the complexity of combining raw data with varying structures, units, and temporal resolutions. This enables the model to process and analyze data from various sources without relying on manual integration schemes. Similar ideas have been proven useful in other domains, too. For instance, \textsf{AtRank} \cite{zhou_atrank_2017} uses attention mechanisms to model user behavior by leaning-based aggregation of heterogeneous data from different user interactions (e.g., items, behaviors, and categories), leading to improved performance in recommendation tasks.
    \item We tackle the challenge of feature engineering by developing shared representations that unify raw data from different sources. This learning-based approach transforms raw features into meaningful representations without losing crucial details, reducing feature engineering efforts. The shared representation also addresses the lack of standardized formats in EHR data, promoting better semantic interoperability and facilitating the integration of external medical knowledge, which enhances the model's generalizability and clinical relevance.
    \item To manage the computational complexity of analyzing massive and heterogeneous EHR data, our method hierarchically aggregates information using three levels of Transformers. This enables the framework to scale efficiently with additional data sources, ensuring robust analysis without compromising performance.
    \item In our experimental results, \textsf{MITST} outperformed the state-of-the-art baseline model \cite{zale_development_2022} for predicting BG levels, demonstrating a 1.7 percentage points (pp) improvement in the average AUROC and a 1.8 pp improvement in the average AUPRC across all classes. The gains were even more pronounced for hypoglycemia (+5.3 pp AUROC, +3.9 pp AUPRC), the most clinically significant category, where accurate predictions are critical for patient outcomes. These findings underscore the superior predictive accuracy and robustness of \textsf{MITST} in managing complex, irregular time-series data from ICU settings.
\end{itemize}

The rest of the paper is organized as follows. In Section \ref{sec:RW}, we review the related work relevant to our research. Section \ref{sec:problem} defines our problem, followed by Section \ref{sec:data}, which presents the data material used in this study. In Section \ref{sec:method}, we describe our proposed method in detail. In Section \ref{sec:results}, we provide our experimental setup and results. We discuss our findings in Section \ref{sec:discussion}, while concluding and outlining future directions in Section \ref{sec:conclusion}.

\section{Related work}
\label{sec:RW}
BG prediction methods can be broadly classified into two categories: physiology-based and data-driven approaches \cite{felizardo_data-based_2021, woldaregay_data-driven_2019, oviedo_review_2017}. Physiology-based methods model BG dynamics using compartments such as meal absorption, insulin activity, exercise, and individual glucose metabolism \cite{felizardo_data-based_2021}. However, our focus is on data-driven approaches. For a detailed review of physiology-based methods, we refer interested readers to \cite{oviedo_review_2017}. 

Data-driven methods for BG prediction fall under the broader category of clinical time-series forecasting. These datasets are often irregularly sampled, making traditional time-series models unsuitable without pre-processing steps like data imputation or window-based aggregation. Imputation-based techniques \cite{liu_handling_2023} aim to convert irregular time-series into regular ones for use with standard forecasting models. For example, Lipton et al. \cite{lipton_learning_2015} applied forward- and back-filling methods on 13 clinical time-series features and used long short-term memory (LSTM) networks to predict diagnoses. Similarly, Harutyunyan et al. \cite{harutyunyan_multitask_2019} combined imputation with indicator variables and evaluated an LSTM on four clinical tasks. Alternatively, window-based approaches aggregate task-specific features from time-series segments. For example, Harutyunyan et al. \cite{harutyunyan_multitask_2019} used windowed features with logistic regression for ICU mortality prediction.

Some studies have attempted to directly model ITS data by learning a continuous-time model from discrete and irregular observations \cite{wang_unsupervised_2014, liu_clinical_2015}. For example, \cite{liu_clinical_2015} models ITS using multiple Gaussian processes (GPs) in the lower level of a hierarchical system and captures the transitions between GPs by using a linear dynamical system. With advancements in deep learning, ITS-focused methods have emerged. Baytas et al. \cite{baytas_patient_2017} proposed Time-Aware LSTM (T-LSTM), which adjusts hidden states based on time gaps between observations. Bai et al. \cite{bai_interpretable_2018} used an attention mechanism to handle irregular time-intervals between clinical events for diagnosis prediction. Tipirneni et al. \cite{tipirneni_self-supervised_2022} introduced STraTS, a Transformer-based framework for ITS, which learns triplet embeddings (feature name, value, timestamp) to predict mortality.

Unlike most existing methods, \textsf{MITST} inherently handles ITS data without requiring time-series-level imputation. Instead, imputation is applied only at the feature level to fill missing values within timestamp-specific vectors, preserving the raw irregularity of the time-series data. Two notable exceptions that also adapt to the irregularity of time intervals directly within their architectures are T-LSTM \cite{baytas_patient_2017} and STraTS \cite{tipirneni_self-supervised_2022}. T-LSTM models patient data as a sequence of discrete (i.e., non-consecutive) visits, requiring larger datasets across multiple patient encounters. In contrast, \textsf{MITST} focuses on data from a single ICU stay, making it applicable to scenarios with fewer data points. STraTS \cite{tipirneni_self-supervised_2022} handles broader data sources by modeling clinical events as triplets. However, while STraTS supports continuous-valued features, it lacks built-in support for sources with categorical features such as diagnosis codes and treatment types, which limits its utility for comprehensive EHR modeling. In contrast, \textsf{MITST} is designed to jointly embed heterogeneous sources—including both numerical and categorical data—with minimal feature engineering. Additionally, STraTS evaluated their model on mortality prediction, which is a simpler task compared to next BG level forecasting. Furthermore, most prior works use a limited number of EHR sources (e.g., diagnosis or procedure codes), whereas \textsf{MITST} integrates a broader range of clinical data (e.g., vital signs, lab results, medications). This allows \textsf{MITST} to determine the most relevant combination of data sources for a specific clinical task, eliminating the need for predefined aggregation common in traditional ITS methods.

It is worth noting that Transformers, though dominant in natural language processing, are less mature in time-series forecasting \cite{zeng_are_2023}. Traditional machine learning models, such as random forests, XGBoost, or ARIMA combined with handcrafted features, remain competitive in this domain. This also holds for large language models (LLMs) in medical applications \cite{tan_are_2024}. Thus, we compare \textsf{MITST} to the state-of-the-art random forest model developed by Zale et al. \cite{zale_development_2022}, which employs carefully designed features for next BG prediction. 

\section{Problem statement}
\label{sec:problem}
In this paper, we address the problem of forecasting clinical outcomes from multi-source and irregular time-series (ITS) by developing a model that effectively handles the complexity of EHR data to generate accurate predictions. While our method is designed to handle any ITS forecasting task, we focus on predicting the next BG level measurement for ICU patients as a case study to show its effectiveness. 

The inputs consist of ITS data from \( M \) distinct sources (e.g., medication records, lab test results, vital sign measurements). Each source provides data recorded at varying, often irregular, intervals. The training dataset consists of several
labeled examples of the form $(\mathbf{x},\mathbf{y})$, where $\mathbf{y} \in \{0, 1\}^C$ is a one-hot encoded label vector representing the class of the next BG measurement. The total number of classes is denoted by \( C \) (i.e., categories for the levels of the next BG measurement - hypoglycemia, hyperglycemia, or euglycemia). Note that the intervals between consecutive BG measurements vary across the dataset, ranging from 5 minutes to 10 hours.

Each data point $\mathbf{x}$ is a collection of ITS data from all sources, collected at irregular timestamps, and is represented as follows.
    \[
    \mathbf{x}=\{\mathbf{x}(m, t_i) \mid m = 1, \ldots, M; \, i = 1, \ldots, T_m\}
    \label{eq:input}
    \tag{1}
    \]
    where:
    \begin{itemize}
        \item \(\mathbf{x}(m, t_i) \in \mathbb{R}^{d_m}\) represents the raw features at the \( i \)-th time point from source \( m \). In general, we model \(\mathbf{x}(m, t_i)\) as a $d_m$-dimensional vector, each dimension representing a feature having either numeric or categorical value. For instance, 
        for a lab source, the vector consists of lab test names as dimensions and the numerical results of the lab tests as values across these dimensions. For some sources, $d_m$ can be one.
        \item \(t_i \in \mathbb{R}^+\) denotes the timestamp for the \( i \)-th time point corresponding to source \( m \) (e.g., the specific time a lab test was performed).
        \item \( T_m\) is the total number of time points corresponding to source \( m \) (e.g., the number of lab tests conducted for a specific patient during their ICU stay).
    \end{itemize}

The specific categorical and numerical features selected for each source are detailed in \hyperref[tab:appendixA2]{Table.~\ref*{tab:appendixA2}} in the appendix. Additionally, the criteria for including records from each source are provided in \hyperref[tab:appendixA3]{Table.~\ref*{tab:appendixA3}}.

At inference time, given a new input, the model outputs a logit vector \(\hat{\mathbf{y}} \in \mathbb{R}^C\), where each element represents the raw, unnormalized score for each class. A softmax function is then applied to produce the final predicted label. The objective is to \textbf{predict the class of the next BG measurement using only the data available up to the current BG measurement.}

\section{Data material}
\label{sec:data}
The eICU Collaborative Research Database (eICU-CRD) \cite{pollard_eicu_2018} is a large, publicly available, multi-center critical care dataset that provides comprehensive anonymized health data for research purposes. It includes information from 200,859 ICU stays across 208 hospitals in the United States, covering admissions from 2014 to 2015. The dataset captures a wide range of clinical variables, including patient demographics, vital signs, laboratory measurements, medications, severity scores (e.g., APACHE), care plans, and treatment details. Out of 31 tables in the eICU database, 11 tables are selected as outlined in \hyperref[tab:appendixA2]{Table.~\ref*{tab:appendixA2}}. Since the goal of this study is to incorporate as much relevant information as possible into the framework, we only exclude tables with low data coverage or those deemed highly irrelevant to the prediction task, allowing the model to decide which features to leverage. Low data coverage in certain tables is primarily due to incomplete or sporadic recording practices across different hospitals, making these tables unsuitable for robust analysis.

\begin{figure*}[!ht]
    \centering
    \includegraphics[width=\linewidth]{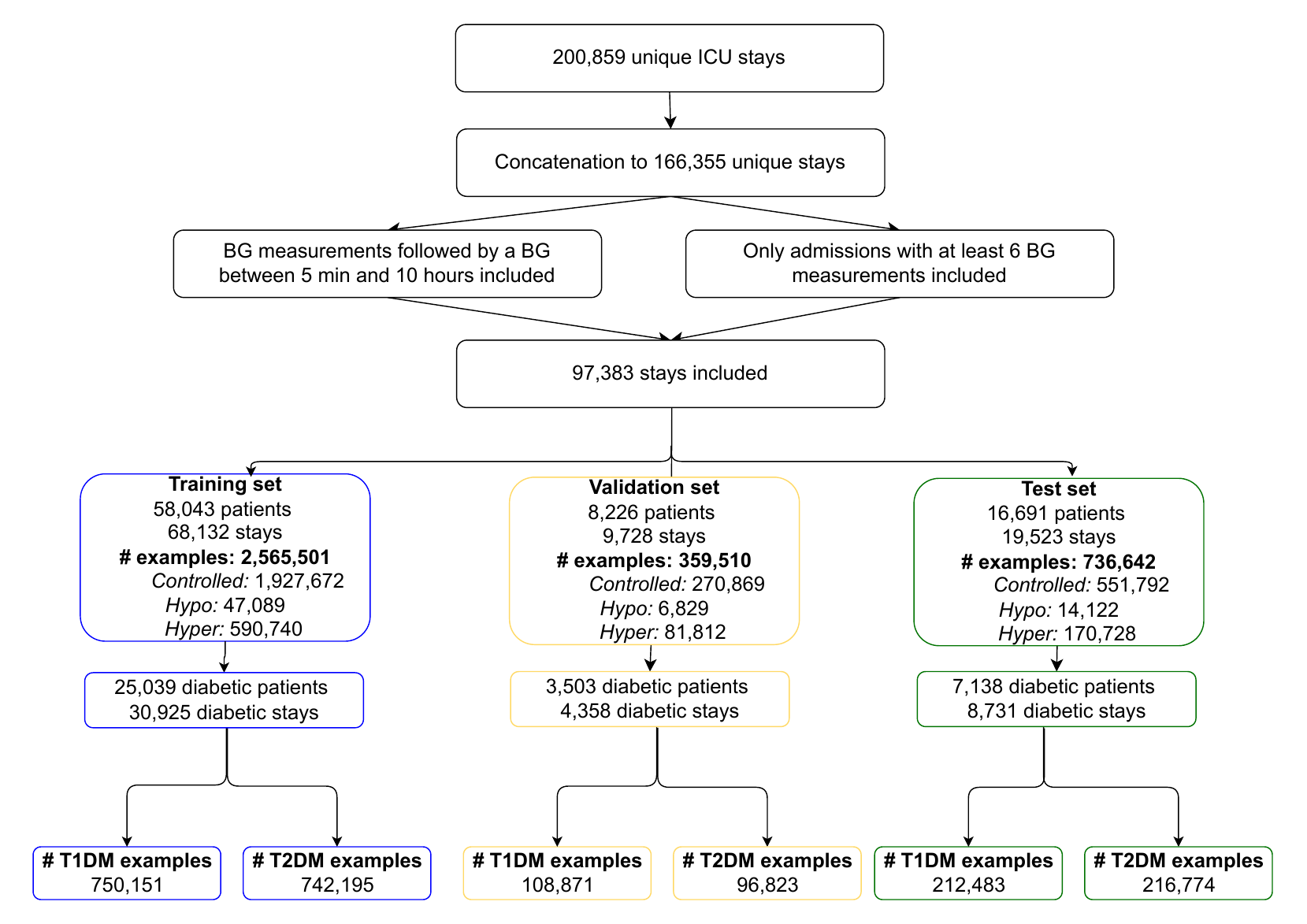}
    \caption{Study flowchart for eICU dataset inclusion criteria and data splits. Each "example" refers to an input-label pair \((\mathbf{x}, \mathbf{y})\), where the input \(\mathbf{x}\) is defined in Eq.~(\ref{eq:input}) and \(\mathbf{y}\) represents the class of the next BG measurement.}
    \label{fig:flowchart}
\end{figure*}

To maintain temporal continuity, we consolidated ICU stays for the same patient only when the admissions occurred at the same hospital and overlapped in time, such as transfers from the hospital floor to ICU. Specifically, we used the ``unitVisitNumber'' column from the eICU \texttt{patient} table to identify sequential visits and retained only those stays that shared identical values for ``hospitalAdmitTime24'', ``hospitalDischargeTime24'', and ``uniquepid'' columns. This ensured that the merged stays reflected a single continuous clinical episode. All timestamps were recalculated relative to the adjusted ``hospitaladmitoffset'', which was reset to zero for each consolidated record. This preprocessing step ensured that the time axis remained consistent and prevented label leakage across merged segments. We include only ICU stays with at least six BG measurements, ensuring that for each input data point, there are at least five prior BG measurements available. Furthermore, each BG measurement must be followed by a subsequent reading within a window of 5 minutes to 10 hours. This interval is chosen in accordance with thresholds established in \cite{zale_development_2022}, ensuring consistency with prior work while allowing our model to operate within clinically relevant prediction windows. The final dataset comprised 97,383 ICU stays, which were split across patients into three non-overlapping sets: 70\% for training, 10\% for validation, and 20\% for testing. \hyperref[fig:flowchart]{Fig.~\ref*{fig:flowchart}} provides a flowchart detailing the inclusion criteria, the number of unique stays, patients, and examples within each set. It also presents the distribution of examples by BG category and includes the statistics of diabetic patients, along with the number of examples for type 1 and type 2 diabetes (T1DM and T2DM, respectively). The classification of T1DM and T2DM was determined using ICD-9 and ICD-10 codes extracted from the \texttt{diagnosis} table. We did our best throughout the paper to adhere to the MI-CLAIM checklist \cite{norgeot_minimum_2020} to ensure the transparency and reproducibility of our study. For further details regarding the dataset and its demographics, refer to the original publication \cite{pollard_eicu_2018}. The details on preprocessing and data imputation can be found in Section \ref{sec:preprocessing}. 

\begin{figure*}[!ht]
    \centering
    \begin{subfigure}{1.0\textwidth}
        \centering
        \includegraphics[width=\linewidth]{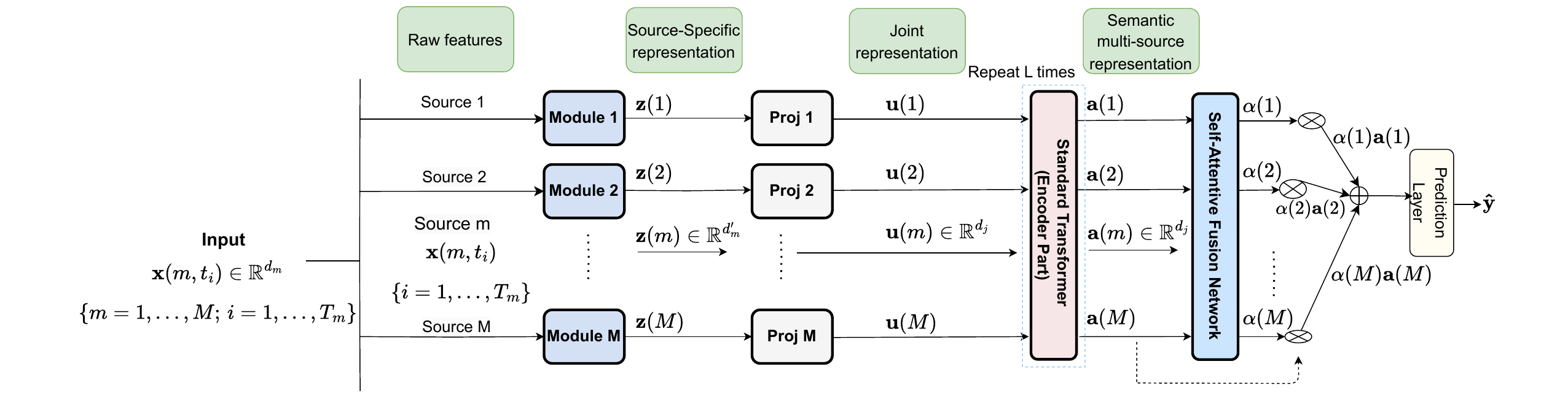}
        \caption{High-level representation of our \textsf{MITST} framework}
        \label{fig:methodA}
    \end{subfigure}
    \vspace{0.5cm}  
    \begin{subfigure}{1.0\textwidth}
        \centering
        \includegraphics[width=\linewidth]{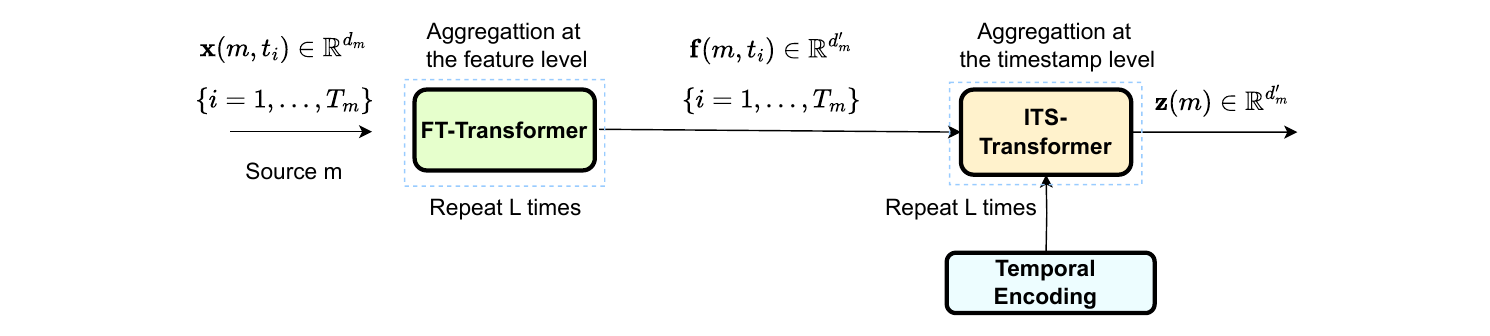}
        \caption{The proposed module for extracting source-specific representation corresponding to source $m$}
        \label{fig:methodB}
    \end{subfigure}
    \caption{The architecture of the proposed \textsf{MITST} method.}
    \label{fig:method}
\end{figure*}

\section{Our method: \textsf{MITST}}
\label{sec:method}
In this section, we present our proposed method, \textsf{MITST}. We detail the key components of the framework, including its modular architecture and integration strategies, which enable effective learning from multi-source medical records.

The time-series for each source $m$ and for every data point $\mathbf{x}$, i.e., \(\mathbf{x}(m, :)\), is first passed through a source-specific module (shown as ``Module 1'' to ``Module $M$'' in \hyperref[fig:methodA]{Fig.~\ref*{fig:methodA}}, respectively), where representations (or embeddings) are extracted based on the characteristics of that particular data source. These source-specific representations are then projected into a shared latent space (via ``Proj 1'' to ``Proj $M$'' in \hyperref[fig:methodA]{Fig.~\ref*{fig:methodA}}) enabling heterogeneous sources to be jointly analyzed despite differences in format and dimensionality.

Next, these unified representations are passed through a Transformer-based encoder architecture (shown as ``Standard Transformer'' in \hyperref[fig:methodA]{Fig.~\ref*{fig:methodA}}), utilizing multi-head attention mechanisms. Each attention head captures distinct inter-source patterns, allowing the model to construct a comprehensive contextual representation. Subsequently, a self-attentive fusion network \cite{lin_structured_2017, liu_attention-based_2022} integrates the source-specific embeddings by adaptively weighting their contributions. This mechanism enables the model to prioritize sources based on their contextual relevance to the prediction task. This final representation is then passed to the prediction layer.

The source-specific module is further elaborated in \hyperref[fig:methodB]{Fig.~\ref*{fig:methodB}}. 
In our overall end-to-end framework, information is hierarchically aggregated using three Transformer-based modules, each operating at a distinct level:

    \begin{itemize}
    \item \textbf{Feature level}: An FT-Transformer \cite{gorishniy_revisiting_2021} (shown in \hyperref[fig:methodB]{Fig.~\ref*{fig:methodB}}) combines categorical and numerical features at each time point within each source. For instance, a single time point from the medication source includes categorical features (e.g., drug name, frequency, route of administration) and numerical features (e.g., dosage). The FT-Transformer generates a unified vector representation for them at each time point, which is subsequently used in the next layers.
    \item \textbf{Timestamp level}: A standard transformer (shown as ``ITS-Transformer'' in \hyperref[fig:methodB]{Fig.~\ref*{fig:methodB}}) aggregates information across irregular timestamps during patient admission. This layer learns interactions between features of a source over time, providing a source-specific representation for each patient admission. Recall that ``ITS'' stands for ``irregular time series''. We refer to this module as the ITS-Transformer to distinguish it from the other modules in two key ways. First, its functionality differs in that it aggregates across time points. Second, it incorporates time encoding (similar to positional encoding in a standard Transformer), unlike the other two layers of aggregation.
    \item \textbf{Source level}: The aggregated representations from all sources are combined using another Transformer (shown as ``Standard Transformer'' in \hyperref[fig:methodB]{Fig.~\ref*{fig:methodB}}), followed by a final self-attention network to make predictions.
    \end{itemize}

The decision to use source-specific representations, followed by their integration into a common framework, is motivated by several factors. Some sources exhibit regular time-series patterns, while others provide only a few time points, if any. Hence, it makes sense to exploit the unique characteristics of each source in designing models that learn representations. Moreover, different sources may vary in their expressive capacities, which in turn influences the size of each source’s representation or embedding—a factor that has been accounted for in our design.

The specifics of each component of the proposed method are described below. The notations used are summarized in \hyperref[tab:notation]{Table.~\ref*{tab:notation}}. An illustrative example of how the proposed model works is provided in \hyperref[fig:AppendixA1]{Fig.~\ref*{fig:AppendixA1}} in the Appendix.

\begin{table*}[htbp]
\centering
\caption{Notations used in this paper.}
\begin{tabularx}{\textwidth}{|c|Y|c|}
\hline
\textbf{Notation} & \textbf{Definition} & \textbf{Dimensionality} \\
\hline
\( M \)            & Number of sources                                                  & -                      \\
\( T_m \)        & Number of time points corresponding to source \(m\)           & -                      \\
\( d_m \)          & Dimensionality of raw features for source \(m\)                    & -                      \\
\( d'_m \)         & Dimensionality of feature embeddings for source \(m\)              & -                      \\
\( d_j \)          & Shared dimensional space size                                      & -                      \\
\( \mathbf{x}(m, t_i) \) & Raw features for the \(i\)-th time point from source \(m\)     & $\mathbb{R}^{d_m}$ \\
\( \mathbf{y} \) & One-hot label vector representing the class label                                   & \( \{0, 1\}^C \)       \\
\( \mathbf{e}_j^{(\text{num})} \) & Embedding for the \(j\)-th numerical feature        & \( \mathbb{R}^{d'_m} \)\\
\( \mathbf{e}_j^{(\text{cat})} \) & Embedding for the \(j\)-th categorical feature      & \( \mathbb{R}^{d'_m} \)\\
\( \mathbf{f}(m, t_i) \) & Feature-aggregated representation for the \(i\)-th time point from source \(m\) & \( \mathbb{R}^{d'_m} \)\\
\( \mathbf{z}(m) \) & Summary of temporal dynamics for source \(m\)                      & \( \mathbb{R}^{d'_m} \)\\
\( \mathbf{u}(m) \) & Projection of source \(m\) into shared dimensional space          & \( \mathbb{R}^{d_j} \) \\
\( \mathbf{a}(m) \) & Final integrated representation for source \(m\)                  & \( \mathbb{R}^{d_j} \) \\
\( \alpha_m \)     & Attention score for source \(m\)                                   & -                      \\
\( \mathbf{v} \)   & Final unified vector representation                                & \( \mathbb{R}^{d_j} \) \\
\( \hat{\mathbf{y}} \) & Logits for prediction                                          & \( \mathbb{R}^C \)     \\
\hline
\end{tabularx}
\label{tab:notation}
\end{table*}

\subsection{Source-specific module}
\sloppy
This module is designed to handle the unique characteristics of data from each source. It aggregates the time-series data \(\mathbf{x}(m, :)\) at two levels -- feature aggregation and timestamp aggregation. Feature aggregation involves combining various features of the data collected at each time point into a single, unified representation. Timestamp aggregation then combines these unified representations across different time points, capturing how the data evolves over time.

\subsubsection{Feature Aggregation (FT-Transformer)} 
Consider source \( m \) with \( n_m \) numerical features and \( c_m \) categorical features, where \( n_m + c_m = d_m \). To aggregate features \( \mathbf{x}(m, t_i) \in \mathbb{R}^{d_m} \) for a single timestamp $t_i$ into a unified representation \( \mathbf{f}(m, t_i) \in \mathbb{R}^{d'_m} \), we employ an FT-Transformer \cite{gorishniy_revisiting_2021}. Although the FT-Transformer was originally designed for tabular data, we adapt it here to handle the mixed types of data (numerical and categorical) within each time point of the irregular time-series. This way, the model can learn an effective representation of the data at each time point.

\textbf{Feature Tokenization:} The FT-Transformer's ``Feature Tokenizer'' component transforms each input feature \( j \) within the vector \(\mathbf{x}(m, t_i)\) into an embedding in \( \mathbb{R}^{d'_m} \). For a numerical feature \( j \), the embedding \( \mathbf{e}_j^{(\text{num})} \) is computed as element-wise multiplication of the feature value \(x_j(m, t_i)\) with a weight vector \( \mathbf{w}_j^{(\text{num})} \in \mathbb{R}^{d'_m} \):
\[
\mathbf{e}_j^{(\text{num})} = b_j + x_j(m, t_i) \cdot \mathbf{w}_j^{(\text{num})}
\tag{2}
\]
where \(b_j\) is the bias term. Also, the embedding for the \( j \)-th categorical feature is computed by a lookup in an embedding table \( \mathbf{w}_j^{(\text{cat})} \in \mathbb{R}^{|\mathcal{C}_j| \times d'_m} \), where \(|\mathcal{C}_j|\) is the number of possible categories for this feature. The embedding for a given category is selected using a one-hot encoded vector \( \boldsymbol{\delta} \) for the corresponding categorical feature:
\[
\mathbf{e}_j^{(\text{cat})} = \boldsymbol{\delta}^T \mathbf{w}_j^{(\text{cat})}
\tag{3}
\]

\textbf{Transformer:} The embeddings for both numerical and categorical features are concatenated with a special [CLS] token, which is often used to capture the overall meaning or summary of the sequence for prediction purposes. The input to the Transformer is then formed as follows:
\[
    \mathbf{f}^{(0)}(m, t_i) = \text{stack}\left[[\text{CLS}], \mathbf{e}_1^{(\text{num})}, \ldots, \mathbf{e}_{n_m}^{(\text{num})}, \mathbf{e}_{n_m+1}^{(\text{cat})}, \ldots, \mathbf{e}_{n_m+c_m}^{(\text{cat})}\right]
\]
These concatenated embeddings are passed through \( L \) layers of the Transformer, with each layer denoted as \( F_l \); \( l = 1, \dots, L \). The output of each layer is obtained recursively:
\[
\mathbf{f}^{(l)}(m, t_i) = F_l(\mathbf{f}^{(l-1)}(m, t_i))
\]
where \(\mathbf{f}^{(l)}(m, t_i)\) represents the embeddings after processing by the \( l \)-th layer of the Transformer. Finally, the embedding corresponding to the [CLS] token after the last Transformer layer is denoted as:
\[
\mathbf{f}(m, t_i) = \mathbf{f}^{(L)}_{\text{[CLS]}}(m, t_i)
\tag{4}
\]
This embedding summarizes the features of the \( i \)-th time point for source \( m \).

\subsubsection{Timestamp Aggregation (ITS-Transformer)} After obtaining the feature-level representations \(\mathbf{f}(m, t_i)\) for each time point \(i\), they are aggregated across time to capture the temporal dynamics within the source \(m\). We employ a standard Transformer that processes the sequence of feature embeddings \(\{\mathbf{f}(m, t_i)\}_{i=1}^{T_m}\). 
The timestamps \( t_i \) are encoded and added to the corresponding feature embeddings \(\mathbf{f}(m, t_i)\) using sinusoidal positional encodings based on their minute-level offset from ICU admission, ensuring that the temporal aspect is fully integrated into the learning process. To handle variable sampling frequencies, we do not resample or aggregate sources into fixed intervals. Instead, we retain the original sampling rates and pass the minute-level timestamps directly into each source-specific transformer. This design preserves temporal granularity, especially for sparse sources such as laboratory tests. Note that no explicit weighting across sources is introduced at this stage.

As in the feature aggregation process, a special [CLS] token is prepended to the sequence of feature embeddings. The sequence is then processed through \( L \) layers of a Transformer, with the [CLS] token capturing the overall temporal information. The final output \(\mathbf{z}(m) \in \mathbb{R}^{d'_m}\), derived from the [CLS] token after \( L \) Transformer layers, serves as a comprehensive summary of the temporal dynamics across all time points for source \( m \).

\subsection{Joint representation}
After we process the data from each source separately, we need to combine these representations into a single form that the model can use to make predictions. To enable this, each source-specific representation \(\{\mathbf{z}(m)\}_{m=1}^{M}\) is projected into a shared space with a fixed number of dimensions (\( d_j \)):
\[
    \mathbf{u}(m) = \text{Linear}_2(\text{GEGLU}(\text{Linear}_1(\text{LayerNorm}(\mathbf{z}(m)))))
    \tag{5}
\]
The input \( \mathbf{z}(m) \) is first normalized with a LayerNorm operation and then passed through a linear transformation \( \text{Linear}_1(d'_m, \,d'_m \, \times \, mult \, \times \, 2) \), where \( mult \) controls the expansion factor of the hidden layer. A GEGLU \cite{shazeer_glu_2020} activation function is applied to introduce non-linearity. Note that the same activation function is also used in all Transformer layers due to its better performance \cite{shazeer_glu_2020}. The result is then projected to the target dimensionality \( d_j \) via a second linear transformation \( \text{Linear}_2(d'_m \, \times \, mult, \, d_j) \). The final output \( \mathbf{u}(m) \) represents the source \( m \) in the shared space. This process is applied independently to each source, yielding the set of shared representations \(\{\mathbf{u}(m)\}_{m=1}^{M}\). This step ensures that the transformed data from each source can be effectively combined and compared in the subsequent steps.

\subsection{Semantic multi-source representation}
After obtaining the shared representations \(\{\mathbf{u}(m)\}_{m=1}^{M}\) for each source, these representations are integrated using a multi-head attention mechanism \cite{vaswani_attention_2017}. Multi-head attention allows the model to look at the data from multiple perspectives at once. Each head in the attention mechanism can focus on different parts of the data or different types of relationships between sources.
By using multiple heads, the model can build a more comprehensive understanding of the data. The outputs from all heads are then concatenated and linearly transformed to produce the final integrated representation \( \mathbf{a}(m) \in \mathbb{R}^{d_j}\). 

\subsection{Self-attentive fusion network}
Once we obtain the multi-source representations, the self-attentive fusion network combines them into a single, unified representation that the model can use to make a final prediction. Given \(\{\mathbf{a}(m)\}_{m=1}^{M}\), we first perform an affine transformation followed by a self-attention mechanism to combine these representations into a unified vector:
\[
\bar{\mathbf{a}}_m = \tanh(\mathbf{w}_f\mathbf{a}(m) + \mathbf{b}_f)
\tag{6}
\]
\[
\alpha_m = \frac{e^{\bar{\mathbf{a}}_m^\top \mathbf{u}_f}}{\sum_m e^{\bar{\mathbf{a}}_m^\top \mathbf{u}_f}}
\tag{7}
\]
\[
\mathbf{v} = \sum_m \alpha_m \mathbf{a}(m)
\tag{8}
\]
Here, \(\mathbf{w}_f \in \mathbb{R}^{d_f \times d_j}\) and \(\mathbf{b}_f \in \mathbb{R}^{d_f}\) are learnable parameters of the affine transformation, and \(\mathbf{u}_f \in \mathbb{R}^{d_f}\) represents the context query vector that is learned during training. The attention scores \(\alpha_m\) reflect the contextual relevance of each source. These scores are used to produce the final representation \(\mathbf{v}\) by taking a weighted sum of the source-specific representations.

This layer learns source-specific attention weights via context-aware interaction between each source embedding and a trainable query vector. This data-driven mechanism adaptively down-weights noisy or redundant signals from densely sampled sources (e.g., vitals) when sparse sources (e.g., labs) provide more informative cues for the prediction task, without requiring handcrafted rules or pre-defined weighting for each source. For example, during hypoglycemia prediction, lab measurements such as blood glucose or serum insulin may receive higher weights than vital signs, which fluctuate more frequently but are less predictive in this context.

\subsection{Prediction layer}
The final unified representation vector \(\mathbf{v}\) is passed through a prediction layer consisting of a LayerNorm operation followed by a ReLU activation function and a fully connected linear layer. The prediction layer computes the logits \(\hat{\mathbf{y}} \in \mathbb{R}^C\), representing the raw, unnormalized predictions for each class. This process is formulated as follows:
\[
\hat{\mathbf{y}} = \text{Linear}_3(\text{ReLU}(\text{LayerNorm}(\mathbf{v})))
\tag{9}
\]
These logits \(\hat{\mathbf{y}}\) are used as the input to the cross-entropy loss function, which is employed to train the \textsf{MITST} architecture.

\section{Experiments}
\label{sec:results}
In this section, we outline the experimental setup used to evaluate the performance of the proposed \textsf{MITST} model for predicting  BG levels in ICU patients. We then present our experimental results, organized around five key research questions: \textbf{Q1}: How does the model's predictive performance compare to the baseline across different BG level classes (Section \ref{sec:Q1})? \textbf{Q2}: When the model makes mistakes, what is the clinical severity of these errors (Section \ref{sec:Q2})? \textbf{Q3}: How does the performance of the model degrade as the time to the next BG measurement increases (Section \ref{sec:Q3})? \textbf{Q4}: How does the model perform across different diabetic patient cohorts (Section \ref{sec:Q4})? \textbf{Q5}: Can the model generalize to a different clinical prediction task, such as in-hospital mortality (Section \ref{sec:Q5})? \textbf{The source code, along with all implementation details, is publicly available at: \url{https://github.com/zavareh89/MITST}}.
\subsection{Experimental setup}
We begin by describing the reference methods used for comparison with our framework, followed by details on preprocessing and data imputation strategies, approaches to handle data imbalance during \textsf{MITST} model training, evaluation criteria employed in this study, and the hyperparameter settings adopted in the training process.
\subsubsection{Competing method}
Since the primary objective of this study is to evaluate the performance of \textsf{MITST} against state-of-the-art clinical models that rely on predefined aggregation and manually engineered features for BG level prediction using the EHR data, we select a recent approach proposed by Zale et al. \cite{zale_development_2022} as the baseline. In their work, the authors implemented several machine learning models, and the best-performing one is a random forest (RF) model. We adopt this RF model as our baseline for comparison. To ensure a consistent and fair comparison, we use the same set of features specified in their study, as long as the features are available in the eICU database. We also apply the same preprocessing and data imputation techniques as described in their methodology. Furthermore, we use the same training, validation, and testing splits for both the baseline and our proposed method. It is important to note that the original paper did not apply any specific techniques for addressing class imbalance in the dataset. Accordingly, we trained the RF model using the entire training set without any re-sampling or class-weighting adjustments.

To further strengthen our comparative analysis, we employed STraTS \cite{tipirneni_self-supervised_2022}—a recent Transformer-based model for irregular clinical time series—as an additional baseline selected from the deep learning-based methods. We carefully adapted the original open-source codebase for the BG prediction task, extracting 102 features from our dataset and formatting them as required triplets (feature name, value, timestamp). Due to STraTS's architectural constraints, only features with continuous values could be incorporated, as its initial embedding layer supports only continuous-valued inputs. This limitation prevented the inclusion of categorical sources such as diagnoses or treatments, thereby reducing the richness of input data. For fairness, we employed the same patient splits, data preprocessing, and class imbalance handling as in our method. The maximum sequence length, which was 880 in the original STraTS implementation to prevent memory overflow, was increased to 1024 observations in our adaptation to better accommodate the high temporal density of our data. Hyperparameters were tuned for our task to ensure optimal performance.

Additionally, as in \cite{zale_development_2022}, we also report the results for a Null model, where the class of the next BG measurement is predicted same as the class of the current BG measurement. This approach is known as "last observation carried forward (LOCF)," a common technique in time-series analysis \cite{moritz_comparison_2015} that assumes the most recent observation is indicative of the immediate future. 

\subsubsection{Preprocessing and data imputation}
\label{sec:preprocessing}
The training dataset includes examples extracted from all patients. All numerical features are standardized using z-score normalization (i.e., centering the features around zero with a unit variance). For features representing measurements across different dimension values (e.g., lab test results corresponding to different test names), the standardization is performed separately for each dimension value. For instance, each unique lab test (e.g., albumin, hemoglobin) is standardized independently to preserve the distributional characteristics of the specific measurement. Missing numerical values are imputed using zeros, which correspond to the mean value after z-score normalization. For categorical features with missing values, a separate category (e.g., "unknown" or "other") is assigned to indicate missing values explicitly. Timestamp information is extracted from the time offset columns available in the eICU tables and is adjusted based on the \texttt{hospitalAdmitOffset} column in the \texttt{patient} table to align all events relative to the patient's admission time. For the medication source, each drug administration record is repeated according to its stated frequency until the specified stop timestamp (i.e., \texttt{drugStopOffset} value). For lab results, outlier values are removed using the 0.05\% and 99.95\% quantiles as thresholds to ensure that extreme values do not distort the analysis.

\subsubsection{Evaluation metrics}
To evaluate the predictive performance, we employ widely used metrics in medical research: Area Under the Receiver Operating Characteristic Curve (AUROC), Area Under the Precision-Recall Curve (AUPRC), positive predictive value (PPV), negative predictive value (NPV), sensitivity (recall), and specificity. We also report prevalence for each class, which refers to the proportion of examples in the dataset belonging to that class. The formulas for each metric are defined as follows:
\begin{align*}
\text{PPV} &= \frac{\text{TP}}{\text{TP} + \text{FP}} \\
\text{NPV} &= \frac{\text{TN}}{\text{TN} + \text{FN}} \\
\text{Sensitivity} &= \frac{\text{TP}}{\text{TP} + \text{FN}} \\
\text{Specificity} &= \frac{\text{TN}}{\text{TN} + \text{FP}}
\end{align*}
Here, TP, FP, TN, and FN refer to the numbers of true positive, false positive, true negative, and false negative predictions, respectively. Also, permutation tests are used to assess the statistical significance of performance differences between models and bootstrapping is used to report 95\% confidence intervals (CI) for AUROC and AUPRC.

\subsubsection{Class imbalance handling}
As shown in \hyperref[fig:flowchart]{Fig.~\ref*{fig:flowchart}}, the distribution of classes is highly imbalanced, with the majority of examples belonging to the euglycemia class. Specifically, the training set contains 1,927,672 euglycemia examples, compared to only 47,089 hypoglycemic and 590,740 hyperglycemic examples. This imbalance poses a challenge for model training, as the model may become biased towards the majority class, leading to poor performance on minority classes \cite{chen_survey_2024, ghosh_class_2024}. To mitigate this issue, we apply uniformly at random undersampling without replacement at the beginning of each training epoch. This procedure ensures that the number of samples per class is equal, thereby balancing the distribution of training examples. Undersampling is a suitable choice in this context because many of the euglycemia class examples are temporally correlated or represent repeated measurements from the same patients. Thus, retaining the entire set would not only increase training time but also introduce redundancy, potentially leading to overfitting, while adding limited new information to the model. The number of training epochs is carefully chosen such that more than 90\% of the original euglycemia class examples are seen at least once by the model during training, ensuring sufficient exposure to the majority class without introducing significant redundancy. It should be noted that this undersampling technique is only applied to the training set. During validation and testing, no resampling is performed, and evaluation metrics are calculated using the entire validation and test sets to accurately reflect model performance across the original, imbalanced class distribution. 

To assess the effect of class imbalance handling on the baseline, we implemented a balanced RF model \cite{chen_using_2004} as a comparative variant. In this approach, each decision tree in the RF ensemble draws an equal number of bootstrap samples from both the minority and majority classes, effectively simulating a per-tree undersampling strategy. This configuration differs from the standard RF model used in the original study \cite{zale_development_2022}, which did not report any explicit balancing method. As shown in \hyperref[tab:appendixA4]{Table.~\ref*{tab:appendixA4}} in the appendix, the balanced RF modestly improved AUROC and sensitivity for the hypoglycemia class but degraded overall macro-averaged performance for these two metrics. Hyperparameters for both RF variants were tuned via cross-validation on the validation set. Based on these findings, we retained the original unbalanced RF setup as the primary baseline in our main results to ensure consistency with prior work and to present the most competitive benchmark.

\subsubsection{Experimental parameter and training settings}
To deal with missing sources for certain data points, we introduce special tokens to represent the absence of data. Specifically, if a source is missing for a particular data point, a single placeholder timestamp is assigned to that source, along with special tokens that indicate the absence of feature values. This approach ensures that the model receives a consistent input format across all data points, allowing it to differentiate between missing and available sources during training and inference. To manage the computational complexity of the ITS-Transformer, we set a maximum sequence length of 512 for each source. For sources with more than 512 time points, only the most recent measurements are retained, and older time points are truncated. This strategy ensures that the ITS-Transformer model focuses on the most relevant and up-to-date information while significantly reducing time complexity. It is worth noting that most sources typically have fewer than 512 time points per example, so this truncation primarily affects high-frequency sources such as vital signs, where continuous monitoring can result in long sequences.

For hyperparameter tuning and early stopping, we use AUROC and AUPRC metrics. The final model is trained for 50 epochs using a batch size of 16. All Transformers have a depth of \(L = 4\) layers, with 8 attention heads per layer, and a dimension of 8 for each head. The shared representation dimension (\(d_j\)) is set to 32, and the learning rate is set to 0.0005 using the Adam optimizer. The dimension of source-specific embeddings (\(d'_m\)) is set based on the characteristics of each source. For sources with a larger number of categories and numerical features, including static, unit\_info, diagnosis, lab, and medication sources, \(d'_m\) is set to 32. For all other sources, \(d'_m\) is set to 16. This approach balances model complexity according to the feature diversity of each source. During training, early stopping is applied to prevent overfitting. Specifically, if the sum of the macro-average AUROC and AUPRC does not improve on the validation set for 5 consecutive epochs, the training process is terminated.

To further examine the robustness of our evaluation, we also conducted a 5-fold patient-level cross-validation using the same features, model configuration, and training procedures. As shown in \hyperref[tab:appendixA5]{Table.~\ref*{tab:appendixA5}} in the appendix, the average performance across folds remained consistent with the original hold-out results—for instance, AUROC and sensitivity for hypoglycemia were 0.917 $\pm$ 0.001 and 0.837 $\pm$ 0.002, closely matching the original split values (0.915 and 0.841, respectively). These results confirm the stability and generalizability of the model's performance across patient subsets. Furthermore, given that the dataset includes approximately 2.6 million training and 0.74 million test examples, the original hold-out split remains clinically realistic. It mimics prospective deployment scenarios, where the model is applied to entirely unseen cohorts.

\subsection{Prediction quality comparison}
\label{sec:Q1}
To evaluate performance in the three-class classification task, we analyze each class independently and the results for the three classes are reported separately. Specifically, binary confusion matrices are used to compare each class against the other two combined (e.g., hyperglycemic vs. euglycemia and hypoglycemic), allowing for the calculation of binary metrics. Although we present all three classes—hypoglycemia, hyperglycemia, and euglycemia glucose levels—for comparison, hypoglycemia and hyperglycemia are of greater clinical significance due to their potential impact on patient outcomes, and will be explored further. For each class, cutpoints used to compute binary classification metrics were selected to maximize the sum of sensitivity and specificity \cite{zale_development_2022}. 

\begin{figure*}[!ht]
    \centering
    \includegraphics[width=\linewidth]{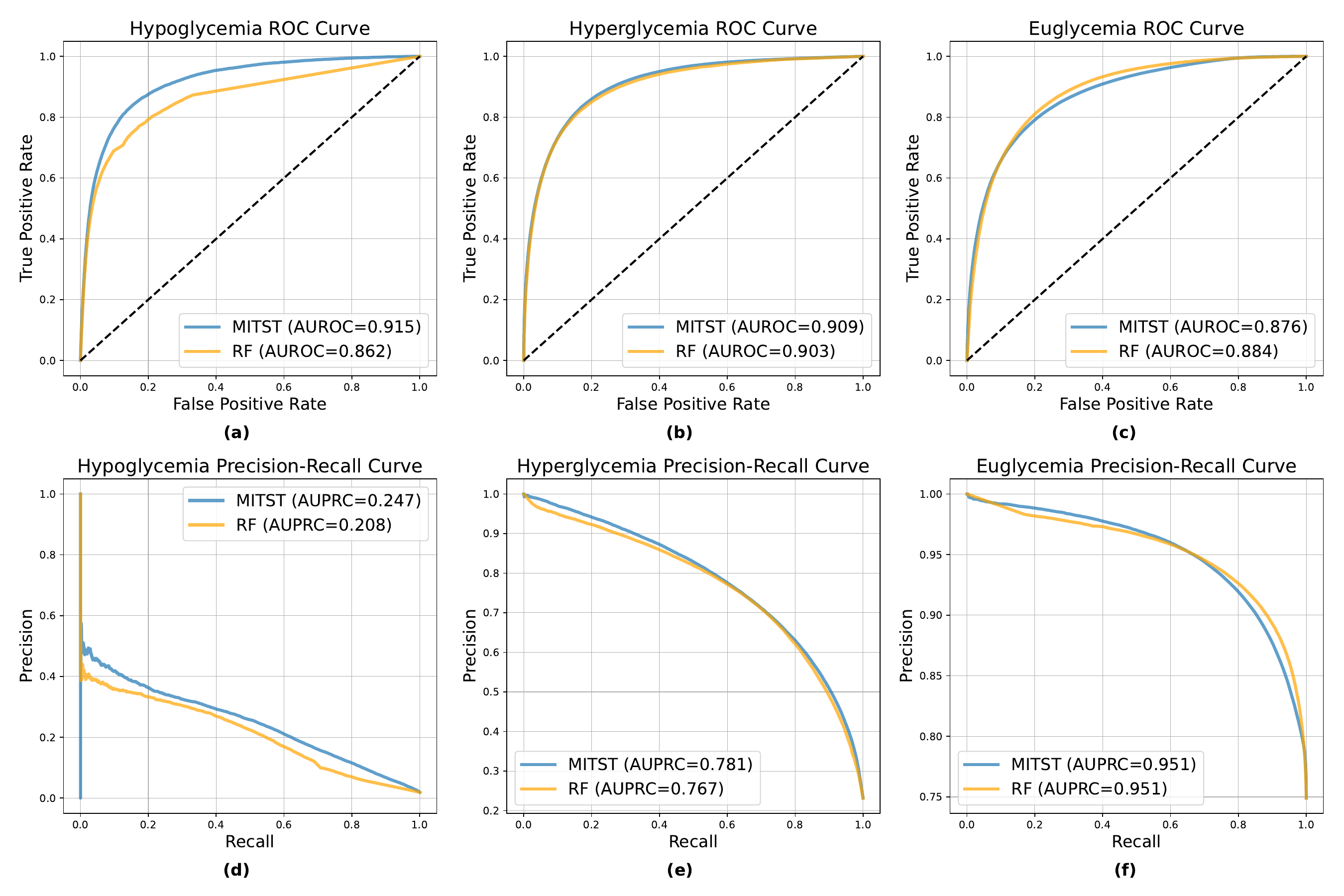}
    \caption{Comparison of ROC and PRC curves for \textsf{MITST} (ours) and Random Forest (RF) \cite{zale_development_2022} models across different glucose level categories (euglycemia, hypoglycemia, and hyperglycemia). Both RF and \textsf{MITST} models were trained using examples drawn from the entire patient cohort.}
    \label{fig:ROCPRC}
\end{figure*}

When comparing the performance of the two models, \textsf{MITST} demonstrates superior capability, particularly in predicting hypoglycemia—a clinically critical but rare event—raising AUROC from 0.862 (CI: 0.858–0.865) to 0.915 (CI: 0.912–0.917) (+5.3 pp) and sensitivity from 0.769 (CI: 0.76.5–0.774) to 0.841 (CI: 0.83.8–0.84.5) (+7.2 pp) as shown in \hyperref[fig:ROCPRC]{Fig.~\ref*{fig:ROCPRC}} and \hyperref[tab:model_comparison]{Table.~\ref*{tab:model_comparison}}. Our test set includes over 14,000 hypoglycemia cases, so an improvement of this magnitude would result in roughly 1,000 additional early detections. In critical care settings, where every minute counts, this could enable a significant clinical improvement of timely interventions, potentially reducing both morbidity and mortality. Similarly, \textsf{MITST}’s AUPRC attains 0.247 (CI: 0.240–0.254), surpassing RF’s 0.208 (CI: 0.201–0.214; \( p < 0.001 \)). In the case of hyperglycemia, although the gap is narrower, \textsf{MITST} still maintains an advantage, with an AUROC of 0.909 (CI: 0.908–0.909) compared to 0.903 (CI: 0.902–0.903) for RF (\( p < 0.001 \)), and an AUPRC of 0.781 (CI: 0.779–0.783), slightly higher than RF’s 0.767 (CI: 0.765–0.769; \( p < 0.001 \)). The lower AUPRC values observed for the minority classes—hypoglycemia and hyperglycemia—are an anticipated outcome of the class imbalance present in the dataset. Since precision-recall curves are more sensitive to class distribution, the smaller number of positive examples in these categories naturally results in lower AUPRC values. Despite this, \textsf{MITST}’s relatively higher AUPRC for these minority classes highlights its robustness in handling imbalanced data. These results demonstrate that the proposed \textsf{MITST} model not only improves predictive performance for minority classes, but also maintains competitive performance for the majority class.

High sensitivity is particularly important in clinical settings, where missing a true positive could have significant consequences. As seen in \hyperref[tab:model_comparison]{Table.~\ref*{tab:model_comparison}}, for hypoglycemia, hyperglycemia, and euglycemia glucose categories, \textsf{MITST} achieves sensitivities of 0.841, 0.833, and 0.778, respectively. In comparison, RF achieves sensitivities of 0.769, 0.818, and 0.816, respectively. These results underscore \textsf{MITST}’s advantage in clinically significant areas, particularly hypoglycemia and hyperglycemia. Moreover, the findings suggest that as the complexity of the prediction task increases—due to the inherent difficulty of the problem and fewer available examples—\textsf{MITST}, a deep learning-based model, excels in its predictive capabilities compared to the traditional RF model. It is worth highlighting that the high PPV and specificity of the Null model for hypoglycemia and hyperglycemia are primarily due to its conservative approach—predicting the next BG level as the same class of the current level. This strategy works well for rare events and stable patients, where BG levels do not fluctuate rapidly. However, while results based on these metrics may appear favorable for the Null model, they mask the Null model’s significant limitations: It lacks sensitivity and fails to capture transitions between BG states, making the Null model useless for proactive clinical intervention.

To compare \textsf{MITST} with more advanced deep learning models, we implemented and evaluated STraTS \cite{tipirneni_self-supervised_2022}, a recent Transformer-based architecture specifically designed for irregular clinical time series. The results are presented in \hyperref[tab:model_comparison_deep]{Table.~\ref*{tab:model_comparison_deep}}, using the same settings as in \hyperref[tab:model_comparison]{Table.~\ref*{tab:model_comparison}}. While STraTS represents a conceptually powerful architecture, its performance in our setting was notably inferior to RF and \textsf{MITST} models. On average, STraTS underperformed even compared to the RF baseline, with a drop of approximately 3.5 pp in AUROC and 5.4 pp in AUPRC across all classes. We attribute this discrepancy to several limitations of STraTS when applied to our task. First, STraTS only supports continuous-valued features and lacks native support for categorical inputs such as diagnosis codes or treatment types, which are critical components of our model’s input space. As a result, a small but clinically informative portion of features was excluded from its representation. Second, its original design and hyperparameter choices were tailored for the simpler task of in-hospital mortality prediction, which involves a coarser temporal resolution. In contrast, BG prediction is more temporally sensitive and clinically nuanced. Despite tuning STraTS for our use case, it remained less effective at capturing the fine-grained, multi-source dependencies essential for accurate BG forecasting.

\begin{table*}[!ht]
    \centering
    \small  
    \caption{Performance metrics comparison across different classes (euglycemia, hypoglycemia, hyperglycemia) and their macro-average values for each model. Both RF and \textsf{MITST} models were trained using examples drawn from the entire patient cohort.}
    \label{tab:model_comparison}
    \resizebox{\textwidth}{!}{  
    \begin{tabular}{|l||c|c|c|c||c|c|c|c||c|c|c|c|c|} 
        \hline
        \multirow{3}{*}{} & \multicolumn{4}{c||}{\textbf{Null model} \cite{zale_development_2022}} & \multicolumn{4}{c||}{\textbf{Random Forest model} \cite{zale_development_2022}} & \multicolumn{4}{c|}{\textbf{\textsf{MITST} model} (ours)} \\ 
        \cline{2-13}
        & \textbf{hypo} & \textbf{hyper} & \textbf{euglycemia} & \textbf{macro avg} & \textbf{hypo} & \textbf{hyper} & \textbf{euglycemia} & \textbf{macro avg} & \textbf{hypo} & \textbf{hyper} & \textbf{euglycemia} & \textbf{macro avg} \\ 
        \hline
        \textbf{prevalence} & 0.019 & 0.232 & 0.749 & -- & 0.019 & 0.232 & 0.749 & -- & 0.019 & 0.232 & 0.749 & -- \\ 
        \midrule
        \multicolumn{13}{l}{\textbf{Metrics}} \\ 
        \hline
        \textbf{AUROC} & --   & --   & --   & --   & 0.862 & 0.903 & 0.884 & 0.883 & 0.915 & 0.909 & 0.876 & 0.900 \\ 
        \hline
        \textbf{AUPRC} & --   & --   & --   & --   & 0.208 & 0.767 & 0.951 & 0.642 & 0.247 & 0.781 & 0.951 & 0.660 \\ 
        \hline
        \textbf{PPV} & 0.298 & 0.698 & 0.892 & 0.629 & 0.081 & 0.602 & 0.922 & 0.535 & 0.096 & 0.596 & 0.926 & 0.539 \\ 
        \hline
        \textbf{NPV} & 0.987 & 0.909 & 0.675 & 0.857 & 0.995 & 0.938 & 0.591 & 0.841 & 0.996 & 0.943 & 0.551 & 0.830 \\ 
        \hline
        \textbf{sensitivity} & 0.312 & 0.698 & 0.891 & 0.634 & 0.769 & 0.818 & 0.816 & 0.801 & 0.841 & 0.833 & 0.778 & 0.817 \\ 
        \hline
        \textbf{specificity} & 0.986 & 0.909 & 0.678 & 0.858 & 0.829 & 0.837 & 0.795 & 0.820 & 0.845 & 0.830 & 0.814 & 0.830 \\ 
        \hline
    \end{tabular}
    }
    \vspace{2mm}  
    \parbox{\textwidth}{\footnotesize \textit{AUROC = Area Under the Receiver Operating Characteristic Curve; AUPRC = Area Under the Precision-Recall Curve; PPV = Positive Predictive Value; NPV = Negative Predictive Value.}}
\end{table*}

\begin{table*}[!ht]
    \centering
    \small  
    \caption{Performance metrics comparison across different classes (euglycemia, hypoglycemia, hyperglycemia) and their macro-average values for STraTS \cite{tipirneni_self-supervised_2022} and \textsf{MITST} models.}
    \label{tab:model_comparison_deep}
    \resizebox{\textwidth}{!}{  
    \begin{tabular}{|l||c|c|c|c||c|c|c|c|c|} 
        \hline
        \multirow{2}{*}{} & \multicolumn{4}{c||}{\textbf{STraTS} \cite{tipirneni_self-supervised_2022}} & \multicolumn{4}{c|}{\textbf{\textsf{MITST} model} (ours)} \\ 
        \cline{2-9}
        & \textbf{hypo} & \textbf{hyper} & \textbf{euglycemia} & \textbf{macro avg} & \textbf{hypo} & \textbf{hyper} & \textbf{euglycemia} & \textbf{macro avg} \\ 
        \hline
        \textbf{prevalence} & 0.019 & 0.232 & 0.749 & -- & 0.019 & 0.232 & 0.749 & -- \\ 
        \midrule
        \multicolumn{9}{l}{\textbf{Metrics}} \\ 
        \hline
        \textbf{AUROC} & 0.872 & 0.845 & 0.827 & 0.848 & 0.915 & 0.909 & 0.876 & 0.900 \\ 
        \hline
        \textbf{AUPRC} & 0.170 & 0.663 & 0.931 & 0.588 & 0.247 & 0.781 & 0.951 & 0.660  \\ 
        \hline
        \textbf{sensitivity} & 0.780 & 0.767 & 0.728 & 0.758 & 0.841 & 0.833 & 0.778 & 0.817 \\ 
        \hline
        \textbf{specificity} & 0.796 & 0.745 & 0.757 & 0.766 & 0.845 & 0.830 & 0.814 & 0.830 \\ 
        \hline
    \end{tabular}
    }
    \vspace{2mm}  
    \parbox{\textwidth}{\footnotesize \textit{AUROC = Area Under the Receiver Operating Characteristic Curve; AUPRC = Area Under the Precision-Recall Curve.}}
\end{table*}

\subsection{Risk analysis comparison}
\label{sec:Q2}
\begin{figure*}[!ht]
    \centering
    \includegraphics[width=\linewidth]{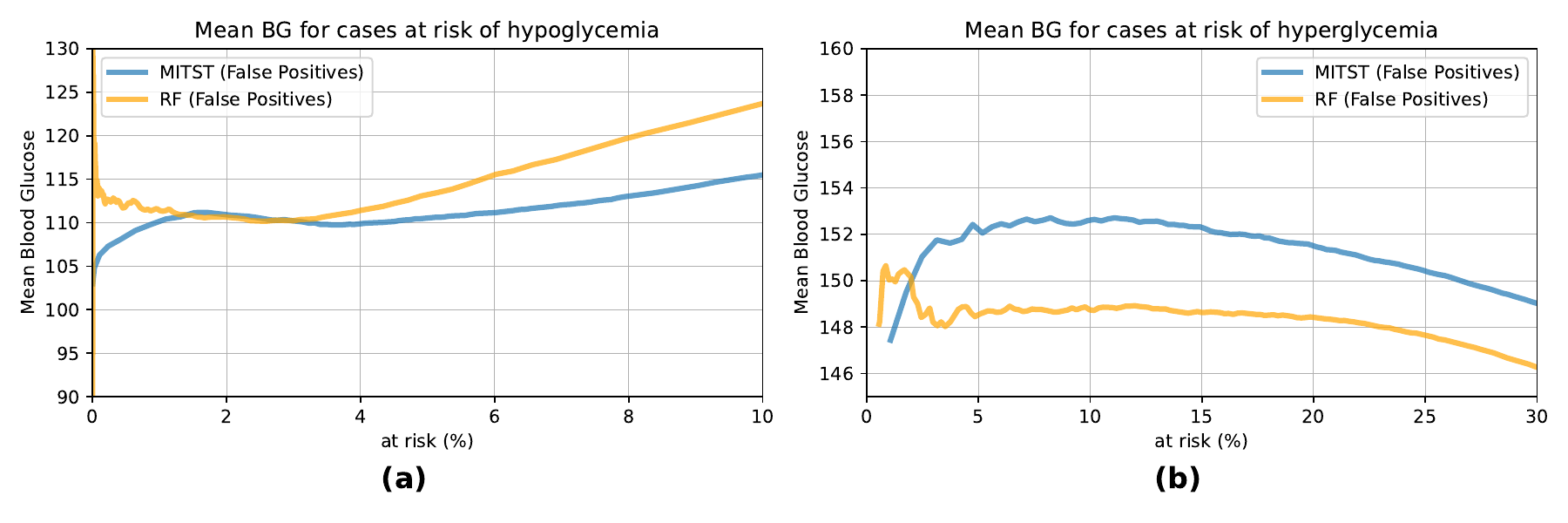}
    \caption{Mean blood glucose for false-positive classifications via RF \cite{zale_development_2022} and \textsf{MITST} (ours) models as a function of the proportion of examples predicted to be at risk of (a) hypoglycemia (b) hyperglycemia.}
    \label{fig:mean_BG}
\end{figure*}

\begin{figure}[!ht]
    \centering
    \includegraphics[width=\linewidth]{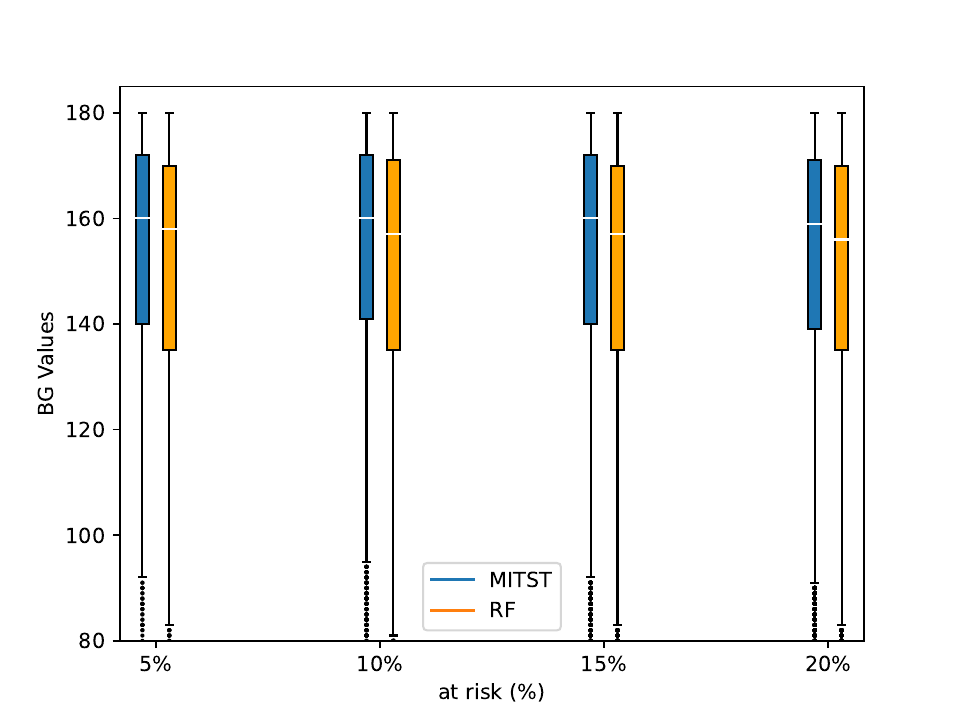}
    \caption{Boxplot of blood glucose values for false positive classifications via RF \cite{zale_development_2022} and \textsf{MITST} (ours) models at selected risk thresholds for hyperglycemia.}
    \label{fig:boxplot}
\end{figure}

While overall performance metrics provide a broad comparison, a more detailed analysis is required to understand the nuances of each model's predictive capabilities for critical conditions (i.e., hypoglycemia and hyperglycemia). To this end, we examine the severity of incorrect predictions made by both models. \hyperref[fig:mean_BG]{Fig.~\ref*{fig:mean_BG}} presents a comparison of the mean BG values for false positive cases identified by the RF and \textsf{MITST} models. The x-axis represents the proportion of examples predicted to be at risk of hypoglycemia or hyperglycemia, while the y-axis shows the corresponding mean BG values. The upper limits of the plots—10\% for hypoglycemia and 30\% for hyperglycemia—are chosen to align with the prevalence of each class in the dataset. In \hyperref[fig:mean_BG]{Fig.~\ref*{fig:mean_BG}}b, higher BG values for false positives in the \textsf{MITST} model compared to the RF suggest that many of these cases may still benefit from clinical intervention, even if they do not strictly meet the hyperglycemia threshold of 180 mg/dL. This is particularly relevant for patients requiring tight glucose control, where maintaining BG levels within a narrower range is critical. For hypoglycemia, as seen in \hyperref[fig:mean_BG]{Fig.~\ref*{fig:mean_BG}}a, the false-positive cases generally deviate further from the clinical threshold of 70 mg/dL. However, \textsf{MITST} demonstrates better performance than RF by predicting values closer to this threshold. To further analyze this effect for hyperglycemia false-positive cases, boxplots of the BG values at selected thresholds are presented in \hyperref[fig:boxplot]{Fig.~\ref*{fig:boxplot}}. These plots clearly show that for the \textsf{MITST} model, the distribution of BG values is more skewed towards the hyperglycemia threshold compared to the RF model. This indicates that \textsf{MITST} not only predicts false positives closer to the clinical threshold but also has a tighter distribution, which can make it more reliable in identifying cases that are closer to requiring intervention.

\begin{figure}[!ht]
    \centering
    \includegraphics[width=0.9\linewidth]{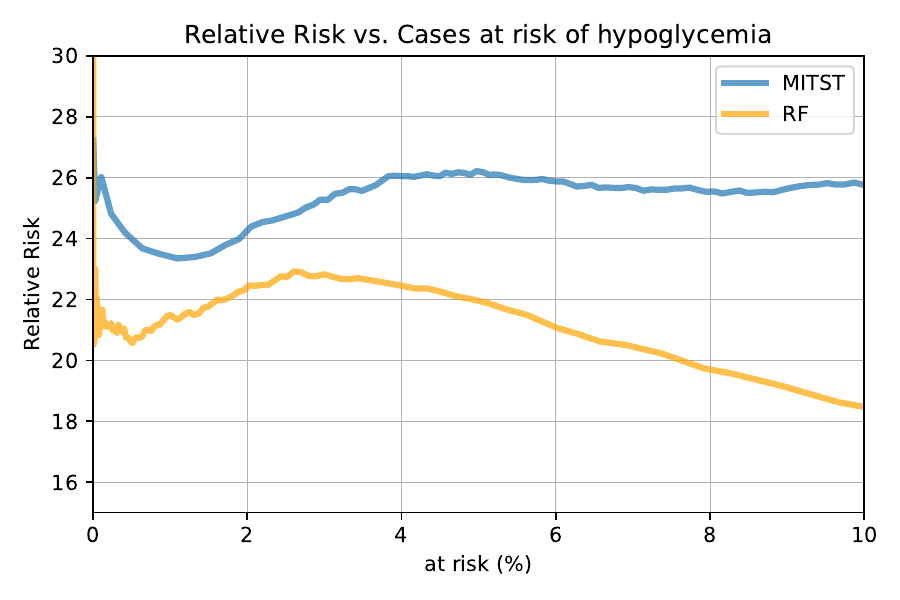}
    \caption{Relative risk of hypoglycemia via RF \cite{zale_development_2022} and \textsf{MITST} (ours) models across percentage of examples predicted to be at risk. The relative risk is calculated as the ratio of the event rate in the predicted hypoglycemia group to the non-hypoglycemia group.}
    \label{fig:RR}
\end{figure}

Given that hypoglycemia is a rare but critical event, we further compare the models by plotting relative risk (RR) for predicted hypoglycemia cases as a function of the percentage of examples predicted to be at risk, as shown in \hyperref[fig:RR]{Fig.~\ref*{fig:RR}}. This metric, defined as the ratio of the likelihood of an event occurring in the predicted group compared to the non-predicted group, provides a clinically intuitive measure of model performance. The plot reveals that \textsf{MITST} consistently maintains a higher relative risk across the range of predicted cases compared to RF, indicating its ability to identify a smaller, high-risk group—an important advantage in clinical decision-making, where reducing unnecessary alarms while capturing high-risk patients is essential.

\subsection{Performance degradation over time}
\label{sec:Q3}
\begin{figure*}[!ht]
    \centering
    \includegraphics[width=0.7\linewidth]{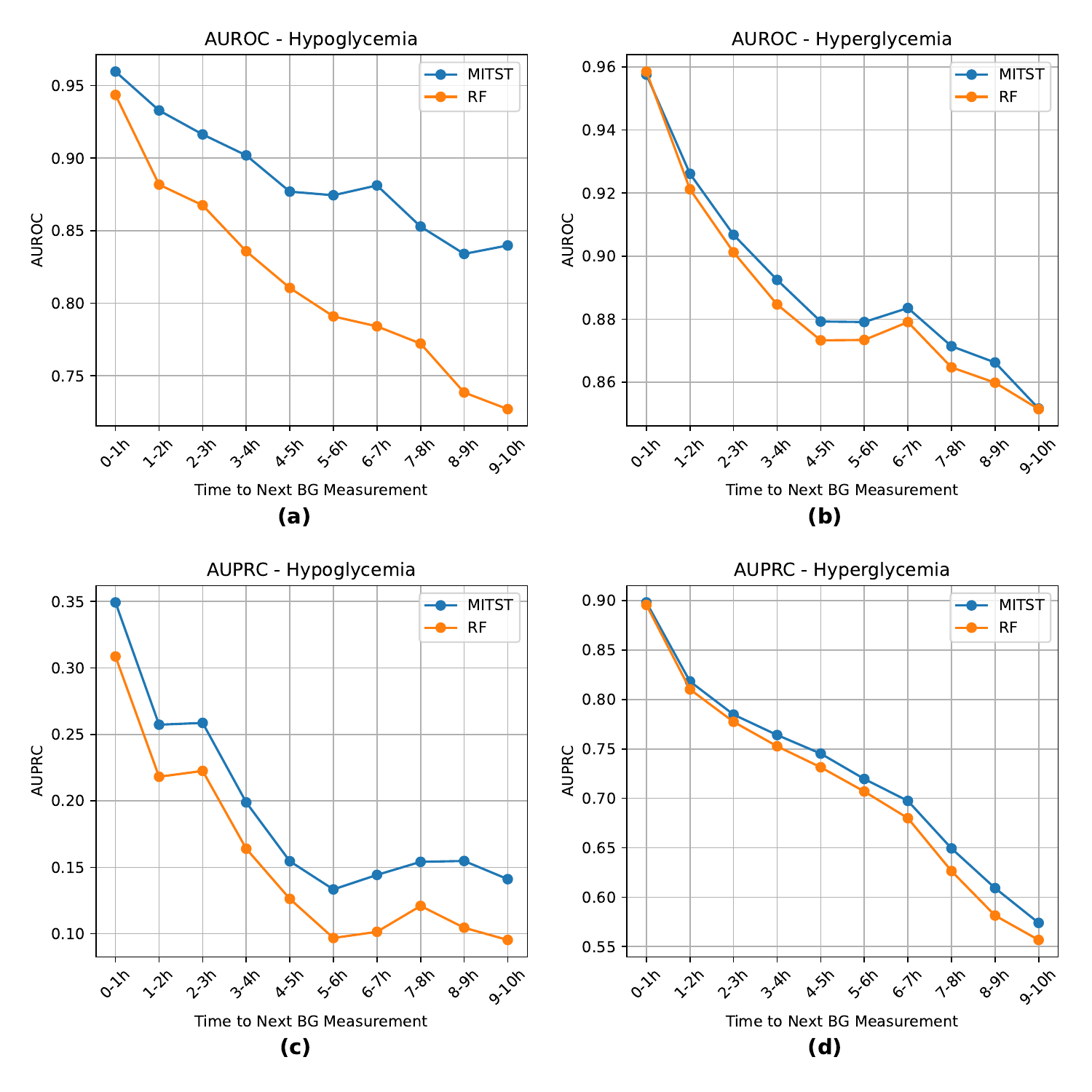}
    \caption{AUROC and AUPRC comparison for hypoglycemia and hyperglycemia classes categorized by the time to the next blood glucose measurement.}
    \label{fig:ROCPRC_time}
\end{figure*}

To evaluate the temporal performance of the models, we analyze their ability to predict BG levels over different time intervals leading up to the next BG measurement. Specifically, we divide the time to the next BG measurement into 1-hour intervals and calculate the AUROC and AUPRC of examples falling within each interval. \hyperref[fig:ROCPRC_time]{Fig.~\ref*{fig:ROCPRC_time}} illustrates how the performance of both \textsf{MITST} and RF models changes as the interval increases. As expected, both models show a decline in performance as the time to the next BG measurement increases, particularly for the hypoglycemia prediction task. This is due to the increased complexity and uncertainty in predicting rare events like hypoglycemia over longer time frames. However, \textsf{MITST} handles these challenging predictions more effectively, showing a slower rate of performance decline compared to RF. The AUROC and AUPRC values for \textsf{MITST} remain consistently higher than RF across all time intervals, with the most pronounced difference observed in the hypoglycemia task. This suggests that \textsf{MITST} is not only better equipped to manage more complex and longer-term predictions, but also maintains a competitive edge over RF in short time intervals, particularly in identifying patients at risk of hypoglycemia.

\subsection{Performance analysis for diabetic cohorts}
\label{sec:Q4}
We also examine how the model’s performance varies between T1DM and T2DM cohorts (defined in Section \ref{sec:data}) for predicting hypoglycemia and hyperglycemia, using the model trained on the overall dataset as discussed in previous sections, along with the combined cohort of patients with diabetes. \hyperref[tab:model_comparison_diabetic]{Table.~\ref*{tab:model_comparison_diabetic}} summarizes the results for these subgroups, highlighting the model's effectiveness across the diabetic populations. The prevalence of hypoglycemia and hyperglycemia events is notably higher in the T1DM cohort compared to the T2DM cohort. This is expected since T1DM patients have a more unstable glucose regulation due to the complete lack of endogenous insulin production, making them more prone to both extremes of glucose levels. In contrast, T2DM patients generally retain some insulin production, leading to relatively less frequent glucose fluctuations, which explains the lower prevalence of these events. Across all diabetic subgroups, our model consistently outperforms the RF model in predicting hypoglycemia and hyperglycemia, demonstrating superior predictive capabilities in all cases. This performance advantage is particularly important for hypoglycemia, where accurate predictions are crucial for patient safety. Additionally, we trained a separate model exclusively on diabetic patients. While the overall performance of both models slightly drop in this setting, the difference between our model and RF remains consistent. 

\begin{table*}[!ht]
    \centering
    \small
    \caption{Performance metrics for RF and \textsf{MITST} models on hypoglycemia and hyperglycemia classes across diabetic cohorts. Both RF and \textsf{MITST} models were trained using examples drawn from the entire patient cohort.}
    \label{tab:model_comparison_diabetic}
    \resizebox{\textwidth}{!}{
    \begin{tabular}{|l||c|c|c|c|c|c||c|c|c|c|c|c|}
        \hline
        \multirow{3}{*}{} & \multicolumn{6}{c||}{\textbf{Random Forest model} \cite{zale_development_2022}} & \multicolumn{6}{c|}{\textbf{\textsf{MITST} model} (ours)} \\ 
        \cline{2-13}
        & \multicolumn{2}{c|}{\textbf{T1DM}} & \multicolumn{2}{c|}{\textbf{T2DM}} & \multicolumn{2}{c||}{\textbf{combined cohort}} & \multicolumn{2}{c|}{\textbf{T1DM}} & \multicolumn{2}{c|}{\textbf{T2DM}} & \multicolumn{2}{c|}{\textbf{combined cohort}} \\ 
        \cline{2-13}
        & \textbf{hypo} & \textbf{hyper} & \textbf{hypo} & \textbf{hyper} & \textbf{hypo} & \textbf{hyper} & \textbf{hypo} & \textbf{hyper} & \textbf{hypo} & \textbf{hyper} & \textbf{hypo} & \textbf{hyper} \\ 
        \hline
        \textbf{prevalence} & 0.028  & 0.369  & 0.017  & 0.299  & 0.022  & 0.328 & 0.028  & 0.369  & 0.017  & 0.299  & 0.022  & 0.328 \\ 
        \midrule
        \multicolumn{13}{l}{\textbf{Metrics}} \\ 
        \hline
        \textbf{AUROC} & 0.847  & 0.876  & 0.858  & 0.88  & 0.854  & 0.879 & 0.899  & 0.878  & 0.915 & 0.888  & 0.907  & 0.884 \\ 
        \hline
        \textbf{AUPRC} & 0.223  & 0.813  & 0.194  & 0.779  & 0.212  & 0.794 & 0.26  & 0.822  & 0.237  & 0.794  & 0.25  & 0.806 \\ 
        \hline
        \textbf{PPV} & 0.092  & 0.638  & 0.07  & 0.61  & 0.083  & 0.622 & 0.108  & 0.639  & 0.088  & 0.60  & 0.10  & 0.617 \\ 
        \hline
        \textbf{NPV} & 0.992  & 0.899  & 0.995  & 0.914  & 0.993  & 0.908 & 0.995  & 0.901  & 0.997  & 0.924  & 0.996  & 0.914 \\ 
        \hline
        \textbf{sensitivity}  & 0.779  & 0.864  & 0.755  & 0.829  & 0.768  & 0.845 & 0.854  & 0.866  & 0.824  & 0.852  & 0.841  & 0.857 \\ 
        \hline
        \textbf{specificity}  & 0.775  & 0.713  & 0.831  & 0.775  & 0.805  & 0.75 & 0.794  & 0.714  & 0.855  & 0.758  & 0.826  & 0.741 \\ 
        \hline
    \end{tabular}
    }
    \vspace{2mm}  
    \parbox{\textwidth}{\footnotesize \textit{AUROC = Area Under the Receiver Operating Characteristic Curve; AUPRC = Area Under the Precision-Recall Curve; PPV = Positive Predictive Value; NPV = Negative Predictive Value.}}
\end{table*}

\subsection{Generalization to mortality prediction task}
\label{sec:Q5}
To assess the generalizability of \textsf{MITST} across tasks, we conducted an additional experiment on in-hospital mortality prediction \cite{safaei_e-catboost_2022}. This setting differs substantially from our original BG prediction task and serves as a strong test for model transferability. We first filtered the dataset to include only ICU stays longer than 24 hours that also contained recorded APACHE IVa scores (described below) \cite{zimmerman_acute_2006}, resulting in a total of 81,184 ICU stays. We used the same train/validation/test split as in the original task, consisting of 56,842, 8,035, and 16,307 stays, respectively. The average in-hospital mortality rate in this subset was 10.1\%. We used the pretrained \textsf{MITST} model from the BG prediction task and fine-tuned it on this new task. To balance transferability and efficiency, we froze all weights except for the fusion layer and the final Transformer module (responsible for semantic multi-source representation). This strategy allows the model to retain the source-specific temporal encoding learned from the original task, while adapting higher-level semantic integration to the new target. We found this configuration to offer a good trade-off between retaining useful representations and allowing task-specific adaptation.

For comparison, we implemented STraTS on this task using the same features and triplet-based input structure as previous task. Unlike \textsf{MITST}, STraTS was trained from scratch rather than fine-tuned from its BG-trained model, due to its poor performance on that task, as shown in Section \ref{sec:Q1}. We used the same hyperparameter settings as those reported in the original STraTS paper for in-hospital mortality prediction. As an additional baseline, we included the APACHE IVa score \cite{zimmerman_acute_2006}, a well-established illness severity scoring system widely used in ICUs for outcome prediction. APACHE IVa incorporates a combination of physiological variables, chronic health conditions, and admission characteristics to estimate in-hospital mortality risk. As discussed in \cite{safaei_e-catboost_2022}, such scoring systems have shown strong predictive performance and remain the clinical standard in many ICUs. It is important to note, however, that APACHE IVa is a proprietary system, and the exact implementation details (e.g., variable mappings, coefficients) are not publicly available. For this reason, we used the APACHE IVa scores directly from the eICU database rather than attempting to replicate the scoring algorithm. Including APACHE IVa as a baseline allows us to benchmark our model against a domain-validated, interpretable alternative.

The comparative results for \textsf{MITST}, STraTS, and APACHE IVa are presented in \hyperref[tab:mortality_comparison]{Table.~\ref*{tab:mortality_comparison}}, using AUROC and AUPRC as the primary evaluation metrics. Despite being fine-tuned on the mortality prediction task with only a subset of its parameters updated (i.e., the fusion layer and final Transformer module), \textsf{MITST} achieved performance comparable to STraTS, which was trained from scratch on this task. This suggests that the source-specific temporal representations learned during pretraining on BG prediction generalized effectively to a different clinical outcome. We note that no additional task-specific hyperparameter optimization was applied to \textsf{MITST}; therefore, its performance could potentially be further improved through targeted tuning. These findings demonstrate the transferability and adaptability of \textsf{MITST} across prediction tasks. In contrast, APACHE IVa outperformed both deep learning models, which can likely be attributed to its highly curated and task-specific input space, optimized specifically for mortality risk estimation in ICUs. By comparison, the feature set used to train \textsf{MITST} and STraTS was originally selected for BG prediction and was intended to be general-purpose. It also lacked several input variables known to be relevant to mortality risk (e.g., detailed ventilator settings, additional lab measurements). Future work could address this limitation by expanding the input space to include a broader range of sources and richer clinical signals per source, potentially narrowing the gap with expert-designed scoring systems.

\begin{table*}[!ht]
    \centering
    \small
    \caption{Comparison of AUROC and AUPRC for in-hospital mortality prediction using three models: APACHE IVa (clinical scoring system) \cite{zimmerman_acute_2006}, STraTS (deep learning baseline)  \cite{tipirneni_self-supervised_2022}, and \textsf{MITST} (fine-tuned from BG prediction task).}
    \label{tab:mortality_comparison}
    \begin{tabular}{|l|c|c|c|}
        \hline
        \textbf{Metric} & \textbf{APACHE IVa \cite{zimmerman_acute_2006}} & \textbf{STraTS  \cite{tipirneni_self-supervised_2022}} & \textbf{\textsf{MITST} (fine-tuned)} \\
        \hline
        AUROC  & 0.824 & 0.804 & 0.806 \\
        AUPRC  & 0.410 & 0.382 & 0.379 \\
        \hline
    \end{tabular}
    \vspace{2mm}
    \parbox{\linewidth}{\footnotesize \textit{AUROC = Area Under the Receiver Operating Characteristic Curve; AUPRC = Area Under the Precision-Recall Curve.}}
\end{table*}

\section{Discussion}
\label{sec:discussion}
In this study, we have developed and validated a novel framework, \textsf{MITST}, for predicting BG levels in ICU patients using a multi-source irregular time-series dataset. Leveraging Transformers, \textsf{MITST} effectively handles the complexity and irregularity of EHR data at multiple levels of granularity, outperforming traditional approaches, particularly in predicting critical conditions such as hypoglycemia and hyperglycemia. \textsf{MITST} showed 1.7 pp improvement in macro-averaged AUROC (\( p < 0.001 \)) and 1.8 pp increase in macro-averaged AUPRC (\( p < 0.001 \)) compared to a state-of-the-art random forest model, significantly enhancing prediction accuracy for at-risk glucose levels. These results emphasize the model’s capacity to provide timely, accurate predictions, enabling clinicians to take proactive measures that improve patient outcomes. Additionally, \textsf{MITST} maintains consistent performance across specific clinical contexts, such as diabetic cohorts. As shown in \hyperref[tab:model_comparison_diabetic]{Table.~\ref*{tab:model_comparison_diabetic}}, the model demonstrated robustness in predicting glucose fluctuations for both type 1 and type 2 diabetes patients. This consistency highlights \textsf{MITST}’s flexibility and scalability by integrating diverse data sources, offering a reliable solution for real-time monitoring and decision support across diverse patient populations in ICU settings, helping mitigate the risk of dysglycemia.

Traditional machine learning models, such as linear regression, random forests, and even more advanced models like recurrent neural networks (RNNs), struggle with irregular time-series data due to their reliance on fixed time intervals and predefined aggregation techniques \cite{morid_time_2023}. In contrast, our Transformer-based approach handles the varying temporal gaps and missing values inherent in the EHR data more effectively through self-attention mechanisms, which captures both short-term and long-term dependencies without a fixed input length. \textsf{MITST}’s learning-based aggregation outperforms traditional predefined aggregation methods by preserving crucial temporal details that would otherwise be lost, particularly for high-risk glucose level predictions. Moreover, unlike RNNs and LSTMs, which may lose contextual relevance over long sequences, \textsf{MITST} can handle complex time-series patterns and varying data intervals, offering superior clinical applicability and interoperability.

One of the key strengths of this study is the use of a large, heterogeneous dataset sourced from multiple hospitals. This enhances the external validity of our findings, ensuring that the model is generalizable across diverse patient populations and clinical settings. However, the reliance on EHR data also presents several challenges. The accuracy and completeness of clinical records can be compromised by incorrect or missing entries, especially those manually recorded by clinicians. For instance, nurse-input errors in medication administration or lab test entries can introduce noise into the dataset, potentially affecting model performance. In clinical practice, this issue could be mitigated by implementing real-time validation checks or flagging potentially erroneous data for review before it is fed into the predictive model. Another advantage of the \textsf{MITST} framework is its ability to dynamically adjust the importance of different data sources based on their relevance to the specific prediction task. For predicting BG levels, the model automatically assigns greater weight to lab results, which are more predictive in this context. Although we have not explicitly tested the model on other clinical tasks, we anticipate that \textsf{MITST} would similarly prioritize relevant sources, such as medication or diagnosis data, when applied to tasks where those inputs are more critical. This adaptability offers a distinct advantage over traditional models, which often rely on manually engineered features tailored to each task. In contrast, \textsf{MITST} eliminates the need for such predefined feature engineering, allowing it to flexibly adapt to a wide range of ICU tasks with minimal adjustment. This dynamic, task-specific source weighting enhances the model’s clinical applicability and generalizability across diverse prediction scenarios in critical care. 

While \textsf{MITST} offers substantial improvements in predicting critical glucose levels, certain limitations must be considered. The complexity of Transformers, which involve multiple layers of self-attention mechanisms, results in longer training times due to the need for large-scale data and extensive computations during model optimization. However, once trained, the model’s inference time is highly efficient—on the order of a few milliseconds—making it well-suited for real-time clinical tasks. The fast inference time mitigates concerns about computational efficiency during real-time deployment, allowing the model to integrate seamlessly into clinical workflows. Although the initial training process is resource-intensive, it is a one-time cost that does not affect the model’s performance or scalability in real-time environments and can be justified by its meaningful impact in detecting rare yet critical conditions like hypoglycemia. Standard models often overlook such infrequent classes due to skewed class distributions, whereas \textsf{MITST} maintains high sensitivity through its end-to-end, attention-based integration of diverse clinical data.

In terms of clinical implications, the \textsf{MITST} framework can help prioritize patients at high or low risk, potentially leading to a reduction in adverse events, such as hypoglycemia, which is associated with increased mortality, morbidity, and length of hospital stay. As shown in \hyperref[fig:mean_BG]{Fig.~\ref*{fig:mean_BG}}, \hyperref[fig:boxplot]{Fig.~\ref*{fig:boxplot}}, and \hyperref[fig:RR]{Fig.~\ref*{fig:RR}}, \textsf{MITST} goes beyond simple classification by identifying individuals at high risk of hypoglycemia and hyperglycemia, offering deeper insights into each patient’s condition. Furthermore, the modular architecture of the proposed model allows for the integration of new data sources with minimal retraining. Only the submodules corresponding to new data sources, along with the top layers of the model (i.e. the layers following the ``Joint Representation'' in \hyperref[fig:methodA]{Fig.~\ref*{fig:methodA}}), require retraining. This design makes \textsf{MITST} a highly dynamic and efficient solution, capable of adjusting to new clinical data streams without the need for a complete model retraining.

Translating predictive models such as \textsf{MITST} into clinical practice demands more than high performance; it requires addressing implementation feasibility, user interaction, and organizational readiness. To demonstrate this, we developed a proof-of-concept web interface that enables real-time interaction with \textsf{MITST}. Users can adjust input features and immediately observe resulting prediction changes, thereby simulating its potential role in a clinical decision support tool (see Appendix).

In typical ICU settings, essential inputs (e.g., heart rate, respiratory rate) are automatically captured by patient monitors and EHR systems, with only a few variables (like diagnoses or medications) needing manual entry—though these, too, can be integrated with EHRs. While the current prototype emphasizes the prediction flow, a complete system would require seamless integration with existing clinical infrastructure to minimize user burden.
Moreover, developing a comprehensive Clinical Decision Support System (CDSS) entails addressing organizational, technical, and regulatory challenges—including workflow integration, coordination among clinicians, IT staff, and administrators, along with patient safety evaluations and usability testing \cite{sutton_overview_2020}. Thus, while promising, successful clinical translation remains a significant future undertaking.

\section{Conclusions and future work}
\label{sec:conclusion}
In this study, we introduced \textsf{MITST}, a novel framework designed to enhance glucose level prediction in ICU patients by leveraging irregular time-series data from multiple clinical sources. By developing a multi-source representation that effectively integrates data from diverse sources such as lab results, medications, and vital signs, our model addresses the inherent complexities of the EHR data, including irregular sampling and heterogeneous data formats. The large-scale dataset used in this study, sourced from multiple hospitals, demonstrates the robustness and generalizability of our model. \textsf{MITST} outperforms a state-of-the-art baseline method, achieving notable improvements in AUROC and AUPRC metrics for clinically significant glucose events. The scalability and flexibility of our approach, which efficiently handles high-dimensional and irregular data, make it a promising tool for real-time decision support in critical care settings. 

While the current study demonstrates the effectiveness of supervised learning in predicting BG levels, future work could explore the application of self-supervised learning techniques. This approach would be particularly useful in scenarios where labeled data is scarce or difficult to obtain. Additionally, future work could explore the use of multi-task learning \cite{chen_m3t-lm_2024} to further enhance \textsf{MITST}'s versatility. Multi-task learning would enable the model to simultaneously learn multiple related tasks—such as predicting other vital signs or clinical outcomes—by sharing representations across tasks, improving generalization and efficiency. Another direction for future work is the application of large language models (LLMs) in EHR data analysis, utilizing their capability to extract insights from unstructured clinical notes and improve multi-source data integration. Also, from an ethical standpoint, it is essential to recognize potential biases within the model due to the underlying dataset, which may not fully represent all ICU patient populations. To mitigate this risk, future studies should prioritize fairness and explainability \cite{payrovnaziri_explainable_2020}, ensuring that the model does not disproportionately benefit or harm any specific subgroup of patients. Finally, a proof-of-concept user interface is developed to demonstrate potential deployment; however, translating this model into a full  Clinical Decision Support System (CDSS) would require significant clinical and organizational integration.



\section*{Declaration of Competing Interest}
The authors declare that they have no known competing financial interests or personal relationships that could have appeared to influence the work reported in this paper.

\section*{Acknowledgements}
Mehdizavareh and Khan were supported by the Novo Nordisk Foundation [grant number NNF22OC0072415]. Cichosz was supported by i-SENS, Inc (Seoul, South Korea).



\clearpage
\appendix
\section*{Appendix}
\setcounter{figure}{0}
\renewcommand{\thefigure}{A\arabic{figure}}

\begin{figure}[h]
    \centering
    \includegraphics[width=\textwidth]{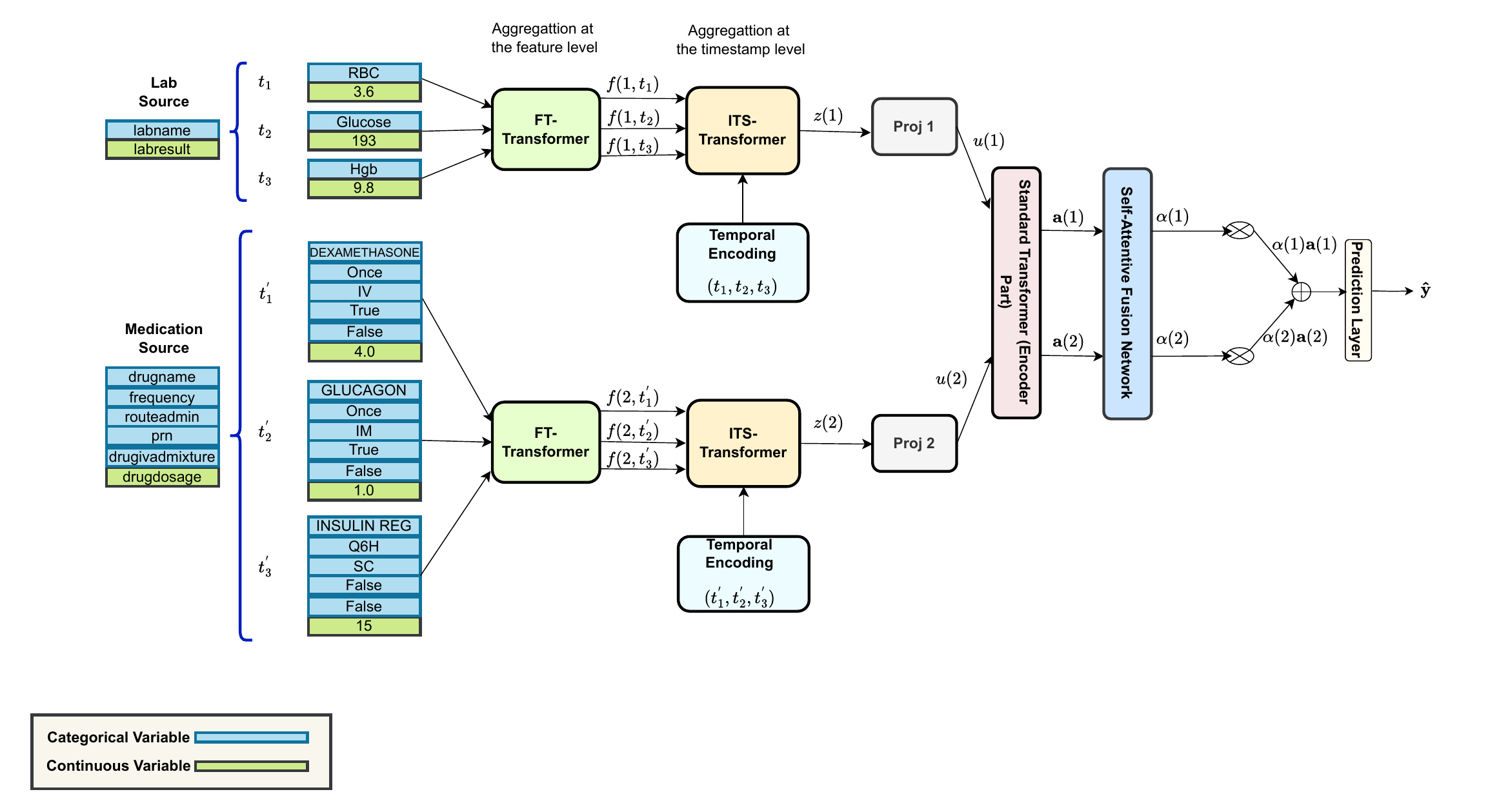}
    \caption{Example of a single input window with data from two sources (lab results and medication), each spanning three time points, fed into the network (\( L = 1\)).}
    \label{fig:AppendixA1}
\end{figure}

\clearpage
\setcounter{table}{0} 
\renewcommand{\thetable}{A\arabic{table}}

\begin{table*}[t]
  \raggedright
  \caption{Description of categorical and numerical features used for each source.}
  \label{tab:appendixA2}
  \begin{tabular}{|l|l|l|p{3.5cm}|p{3.5cm}|}
    \hline
    \textbf{Source index} & \textbf{Source name} & \textbf{eICU table name(s)} & \textbf{Categorical features} & \textbf{Numerical features} \\
    \hline
    1 & Static & \texttt{patient}, \texttt{hospital} & gender, ethnicity, hospital\_id, hospital\_numbeds, hospital\_region, hospital\_admitsource, hospital\_teachingstatus, & weight, age, height \\
    \hline
    2 & Unit Info & \texttt{patient} & unit\_type, unit\_staytype, unit\_admitsource & -- \\
    \hline
    3 & AdmissionDx & \texttt{admissiondx} & admission\_diagnosis & -- \\
    \hline
    4 & Diagnosis & \texttt{diagnosis} & diagnosis, diagnosispriority & -- \\
    \hline
    5 & Lab & \texttt{lab} & labname & labresult \\
    \hline
    6 & IntakeOutput (IO) & \texttt{intakeOutput} & celllabel & cellvalue \\
    \hline
    7 & IO\_total & \texttt{intakeOutput} & -- & num\_registrations, intake, output, dialysis) \\
    \hline
    8 & Past History & \texttt{pastHistory} & pasthistoryvalue & -- \\
    \hline
    9 & Treatment & \texttt{treatment} & treatmentstring & -- \\
    \hline
    10 & Medication & \texttt{medication} & drugname, drugadmitfrequency, drugnotetype, rxincluded, writtenineicu & drugdosage \\
    \hline
    11 & Infusion & \texttt{infusionDrug} & drugname & drugrate \\
    \hline
    12 & GCS Score & \texttt{nurseCharting} & score & -- \\
    \hline
    13 & Sedation Score & \texttt{nurseCharting} & score & -- \\
    \hline
    14 & HR & \texttt{nurseCharting} & -- & value \\
    \hline
    15 & RR & \texttt{nurseCharting} & -- & value \\
    \hline
    16 & SpO2 & \texttt{nurseCharting} & -- & value \\
    \hline
    17 & Temperature & \texttt{nurseCharting} & location & value \\
    \hline
    18 & nibp\_mean & \texttt{nurseCharting} & -- & value \\
    \hline
    19 & ibp\_mean & \texttt{nurseCharting} & -- & value \\
    \hline
    20 & nibp\_systolic & \texttt{nurseCharting} & -- & value \\
    \hline
    21 & ibp\_systolic & \texttt{nurseCharting} & -- & value \\
    \hline
    22 & nibp\_diastolic & \texttt{nurseCharting} & -- & value \\
    \hline
    23 & ibp\_diastolic & \texttt{nurseCharting} & -- & value \\
    \hline
  \end{tabular}
\end{table*}

\clearpage
\begin{table*}[t]
  \raggedright
  \caption{Inclusion criteria for sources used in the study.}
  \label{tab:appendixA3}
  \begin{tabular}{|l|p{10cm}|}
    \hline
    \textbf{Source name} & \textbf{Inclusion criteria} \\
    \hline
    Diagnosis & "diagnosis" frequency > 50 and "diagnosis" column contains one of the following terms: \textit{“diabetes”, “hyperglycemia”, “hypoglycemia”, “glucose”, “insulin”, “kidney”, “pancrea”, “sepsis”, “liver”, “congestive heart failure”}. \\
    \hline
    AdmissionDx & Same criteria as Diagnosis. \\
    \hline
    Lab & "labname" column must be one of the following: \textit{“ALT (SGPT)”, “AST (SGOT)”, “alkaline phos.”, “direct bilirubin”, “total bilirubin”, “BUN”, “creatinine”, “24 h urine protein”, “urinary creatinine”, “sodium”, “potassium”, “chloride”, “magnesium”, “calcium”, “phosphate”, “HDL”, “LDL”, “triglycerides”, “total cholesterol”, “CRP”, “ESR”, “urinary osmolality”, “urinary sodium”, “urinary specific gravity”, “WBC x 1000”, “-lymphs”, “-monos”, “-eos”, “-basos”, “-bands”, “-polys”, “RBC”, “Hgb”, “Hct”, “platelets x 1000”, “MCV”, “MCH”, “MCHC”, “RDW”, “albumin”, “anion gap”, “bicarbonate”, “HCO3”, “Base Deficit”, “Base Excess”, “PTT”, “PT - INR”, “PT”, “pH”, “lactate”, “total protein”, “folate”, “LDH”, “Ferritin”, “bedside glucose”, “glucose”}. \\
    \hline
    Medication & "drugname" frequency > 50 and "drugname" column contains one of the following terms: \textit{“insulin”, “regular”, “lispro”, “aspart”, “glargine”, “detemir”, “humalog”, “novolog”, “lantus”, “metformin”, “glipizide”, “glyburide”, “glimepiride”, “pioglitazone”, “rosiglitazone”, “acarbose”, “miglitol”, “sitagliptin”, “saxagliptin”, “linagliptin”, “exenatide”, “liraglutide”, “levemir”, “dapagliflozin”, “canagliflozin”, “empagliflozin”, “prednisone”, “dexamethasone”, “Decadron”, “hydrocortisone”, “methylprednisolone”, “medrol”, “solumedrol”, “beta-blockers”, “thiazide diuretics”, “niacin”, “atypical antipsychotics”, “statins”, “protease inhibitors”, “pentamidine”, “glucagon”, “quinolones”, “corticosteroids”, “dext” (dextrose), “D5”, “D10” , “D50”, “glucose”}. \\
    \hline
    Infusion & "drugname" column contains one of the following terms: \textit{“insulin”, “dextrose”, “D5”, “D10”, “D50”}. \\
    \hline
    IntakeOutput (IO) & "Celllabel" column contains one of the following terms: \textit{“insulin”, “TPN”, “dextrose”, “D5”, “D10”, “albumin”, “hypertonic”, “NS”, “saline”, “lactated”, “LR”, “NaCl”, “sodium chloride”, “CRRT”, “Urin”, “Urethral”, “Foley”, “dialysis”}. \\
    \hline
    Past History &  "pastHistoryPath" column contains one of the following terms: \textit{“diabetes”, “insulin”, “hypoglycemia”, “hyperglycemia”, “thyroid”, “hypertension”, “CHF”, “renal”, “liver”, “pancreas”, “cirrhosis”}. \\
    \hline
    Treatment & "treatmentstring" column contains one of the following terms: \textit{“insulin”, “D5”, “D10”, “D50”, “oral hypoglycemic administration”, “dialysis”, “glucose”, “nutrition”}. \\
    \hline
  \end{tabular}
\end{table*}

\clearpage
\begin{table*}[!ht]
    \centering
    \small  
    \caption{Comparison of performance metrics for the original and balanced Random Forest (RF) models across glycemic classes (hypoglycemia, hyperglycemia, euglycemia), including macro-averaged values. The original RF model \cite{zale_development_2022} does not apply any explicit class balancing strategy, whereas the balanced RF variant uses per-tree class-balanced bootstrap sampling—drawing equal numbers of samples from both minority and majority classes in each decision tree. This approach is analogous to the undersampling technique used in our deep model (\textsf{MITST}). While the balanced RF slightly improves AUROC and sensitivity for the hypoglycemia class, it results in decreased macro-averaged performance across most metrics. Hyperparameters for both RF models were tuned via cross-validation on the validation set.}
    \label{tab:appendixA4}
    \resizebox{\textwidth}{!}{  
    \begin{tabular}{|l||c|c|c|c||c|c|c|c|c|} 
        \hline
        \multirow{2}{*}{} & \multicolumn{4}{c||}{\textbf{Original RF model} \cite{zale_development_2022}} & \multicolumn{4}{c|}{\textbf{Balanced RF model}} \\ 
        \cline{2-9}
        & \textbf{hypo} & \textbf{hyper} & \textbf{euglycemia} & \textbf{macro avg} & \textbf{hypo} & \textbf{hyper} & \textbf{euglycemia} & \textbf{macro avg} \\ 
        \hline
        \textbf{prevalence} & 0.019 & 0.232 & 0.749 & -- & 0.019 & 0.232 & 0.749 & -- \\ 
        \midrule
        \multicolumn{9}{l}{\textbf{Metrics}} \\ 
        \hline
        \textbf{AUROC} & 0.862 & 0.903 & 0.884 & 0.883 & 0.872 & 0.898 & 0.866 & 0.878 \\ 
        \hline
        \textbf{AUPRC} & 0.208 & 0.767 & 0.951 & 0.642 & 0.204 & 0.746 & 0.946 & 0.632 \\ 
        \hline
        \textbf{PPV} & 0.081 & 0.602 & 0.922 & 0.535 & 0.090 & 0.589 & 0.921 & 0.533 \\ 
        \hline
        \textbf{NPV} & 0.995 & 0.938 & 0.591 & 0.841 & 0.995 & 0.940 & 0.548 & 0.828 \\ 
        \hline
        \textbf{sensitivity} & 0.769 & 0.818 & 0.816 & 0.801 & 0.778 & 0.824 & 0.779 & 0.794 \\ 
        \hline
        \textbf{specificity} & 0.829 & 0.837 & 0.795 & 0.820 & 0.847 & 0.826 & 0.801 & 0.825 \\ 
        \hline
    \end{tabular}
    }
    \vspace{2mm}  
    \parbox{\textwidth}{\footnotesize \textit{AUROC = Area Under the Receiver Operating Characteristic Curve; AUPRC = Area Under the Precision-Recall Curve; PPV = Positive Predictive Value; NPV = Negative Predictive Value.}}
\end{table*}

\clearpage
\begin{table*}[!ht]
    \centering
    \small  
    \caption{Comparison of \textsf{MITST} model performance using (left) the original patient-level train/validation/test split from the main paper and (right) a 5-fold patient-level cross-validation. Metrics are reported per glycemic class (hypoglycemia, hyperglycemia, euglycemia) and as macro-averages. In the cross-validation setting, the reported values represent the mean and standard deviation across folds. Note that for cross-validation, we used a different random partitioning of patients than the original hold-out split used. The results demonstrate strong agreement between the two evaluation strategies.}
    \label{tab:appendixA5}
    \resizebox{\textwidth}{!}{  
    \begin{tabular}{|l||c|c|c|c||c|c|c|c|c|} 
        \hline
        \multirow{2}{*}{} & \multicolumn{4}{c||}{\textbf{\textsf{MITST} (original train/val/test split used in main paper)}} & \multicolumn{4}{c|}{\textbf{\textsf{MITST} (5-fold patient-level cross-validation)}} \\ 
        \cline{2-9}
        & \textbf{hypo} & \textbf{hyper} & \textbf{euglycemia} & \textbf{macro avg} & \textbf{hypo} & \textbf{hyper} & \textbf{euglycemia} & \textbf{macro avg} \\ 
        \hline
        \textbf{prevalence} & 0.019 & 0.232 & 0.749 & -- & 0.019 & 0.232 & 0.749 & -- \\ 
        \midrule
        \multicolumn{9}{l}{\textbf{Metrics}} \\ 
        \hline
        \textbf{AUROC} & 0.915 & 0.909 & 0.876 & 0.900 & 0.917 $\pm$ 0.001 & 0.909 $\pm$ 0.001 & 0.878 $\pm$ 0.001 & 0.901 $\pm$ 0.001 \\ 
        \hline
        \textbf{AUPRC} & 0.247 & 0.781 & 0.951 & 0.660 & 0.251 $\pm$ 0.003 & 0.780 $\pm$ 0.001 & 0.952 $\pm$ 0.001 & 0.661 $\pm$ 0.001 \\ 
        \hline
        \textbf{PPV} & 0.096 & 0.596 & 0.926 & 0.539 & 0.098 $\pm$ 0.001 & 0.592 $\pm$ 0.002 & 0.927 $\pm$ 0.001 & 0.539 $\pm$ 0.001 \\ 
        \hline
        \textbf{NPV} & 0.996 & 0.943 & 0.551 & 0.830 & 0.996 $\pm$ 0.001 & 0.944 $\pm$ 0.001 & 0.550 $\pm$ 0.005 & 0.830 $\pm$ 0.002 \\ 
        \hline
        \textbf{sensitivity} & 0.841 & 0.833 & 0.778 & 0.817 & 0.837 $\pm$ 0.002 & 0.834 $\pm$ 0.001 & 0.779 $\pm$ 0.004 & 0.817 $\pm$ 0.002 \\ 
        \hline
        \textbf{specificity} & 0.845 & 0.830 & 0.814 & 0.830 & 0.852 $\pm$ 0.002 & 0.828 $\pm$ 0.002 & 0.814 $\pm$ 0.003 & 0.831 $\pm$ 0.001 \\ 
        \hline
    \end{tabular}
    }
    \vspace{2mm}  
    \parbox{\textwidth}{\footnotesize \textit{AUROC = Area Under the Receiver Operating Characteristic Curve; AUPRC = Area Under the Precision-Recall Curve; PPV = Positive Predictive Value; NPV = Negative Predictive Value.}}
\end{table*}

\clearpage
\section*{Supplementary Material: Interactive \textsf{MITST} Demo System}
\renewcommand{\thefigure}{S\arabic{figure}}
\setcounter{figure}{0}

We developed a web-based proof-of-concept interface that executes the pre-trained \textsf{MITST} model on custom input data. Hosted on an Aalborg University server, this interface simulates clinical inference for a hypothetical ICU patient. Note that it is intended solely for demonstration purposes and not for production deployment. A live demo is available at http://130.225.37.30:8501/.

\subsection*{System Overview and Usage}
\begin{figure*}[h]
    \centering
    \includegraphics[width=\textwidth]{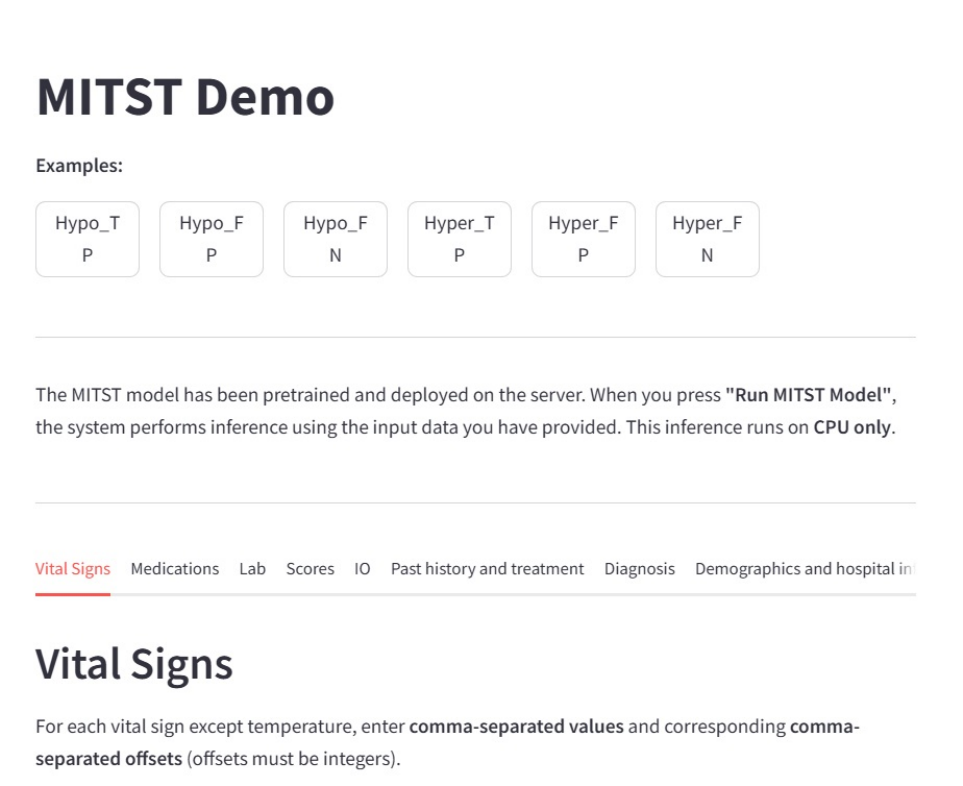}
    \caption{Main interface of the \textsf{MITST} demo system. Users can choose pre-filled examples or manually navigate between tabs to provide inputs. Pressing ``Run \textsf{MITST} Model'' button triggers inference.}
    \label{fig:main_ui}
\end{figure*}

Figure~\ref{fig:main_ui} illustrates the main interface of the system, which is organized into multiple tabs representing different data sources. After entering or loading the required data, users can press the \textbf{``Run \textsf{MITST} Model''} button to obtain a predicted glucose level label (i.e., hypoglycemia, euglycemia, or hyperglycemia). The inference is performed on the server using a pretrained model.

To simplify usage, we provide \textbf{six pre-filled example templates} from the test set of the eICU database, accessible through dedicated buttons in the interface (see Figure~\ref{fig:main_ui}). These include:

\begin{itemize}
    \item 3 examples for hypoglycemia: True Positive, False Positive, False Negative.
    \item 3 examples for hyperglycemia: True Positive, False Positive, False Negative.
\end{itemize}

\begin{figure*}[h]
    \centering
    \includegraphics[width=\textwidth]{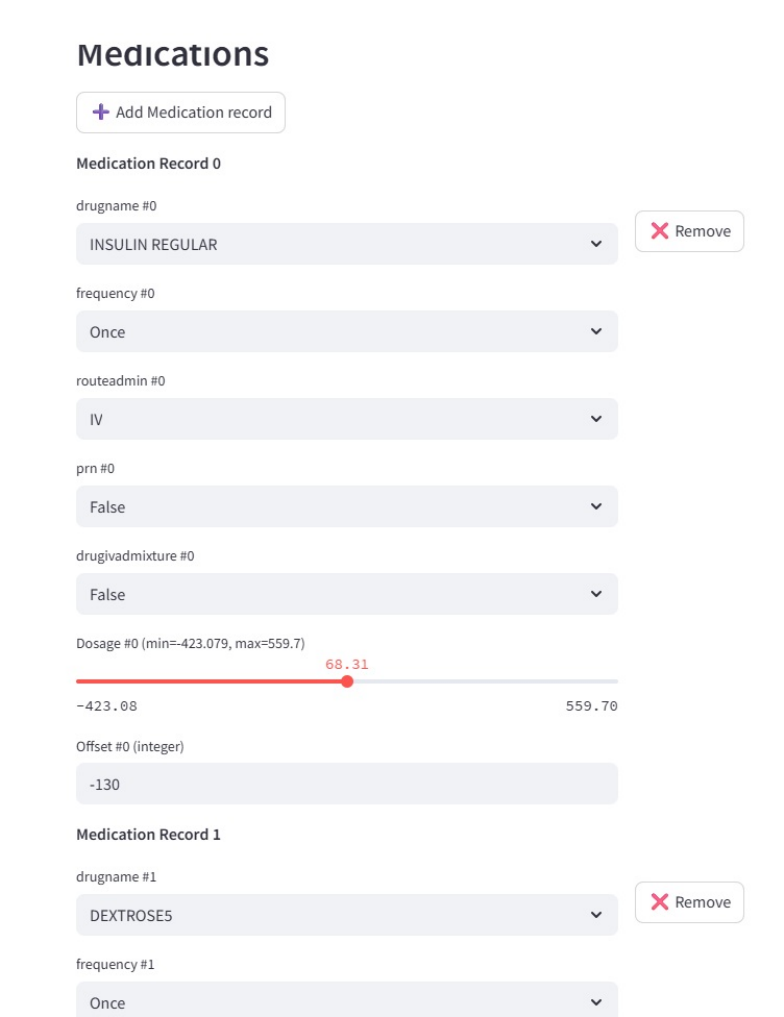}
    \caption{Example of a pre-filled case for hypoglycemia (False Positive). Users can modify any value to test sensitivity of the model to input changes.}
    \label{fig:prefilled_example}
\end{figure*}

Selecting any of these automatically fills all input tabs with the corresponding data (Figure~\ref{fig:prefilled_example}). Users can modify any value and re-run the model to observe how predictions change in real-time.

\subsection*{Input Structure and Tabs}
The system handles heterogeneous clinical data following the \textsf{MITST} training format. Input is divided across multiple tabs—each corresponding to one or more data sources—to streamline user interaction and visualization.

\begin{figure*}[h]
    \centering
    \includegraphics[width=\textwidth]{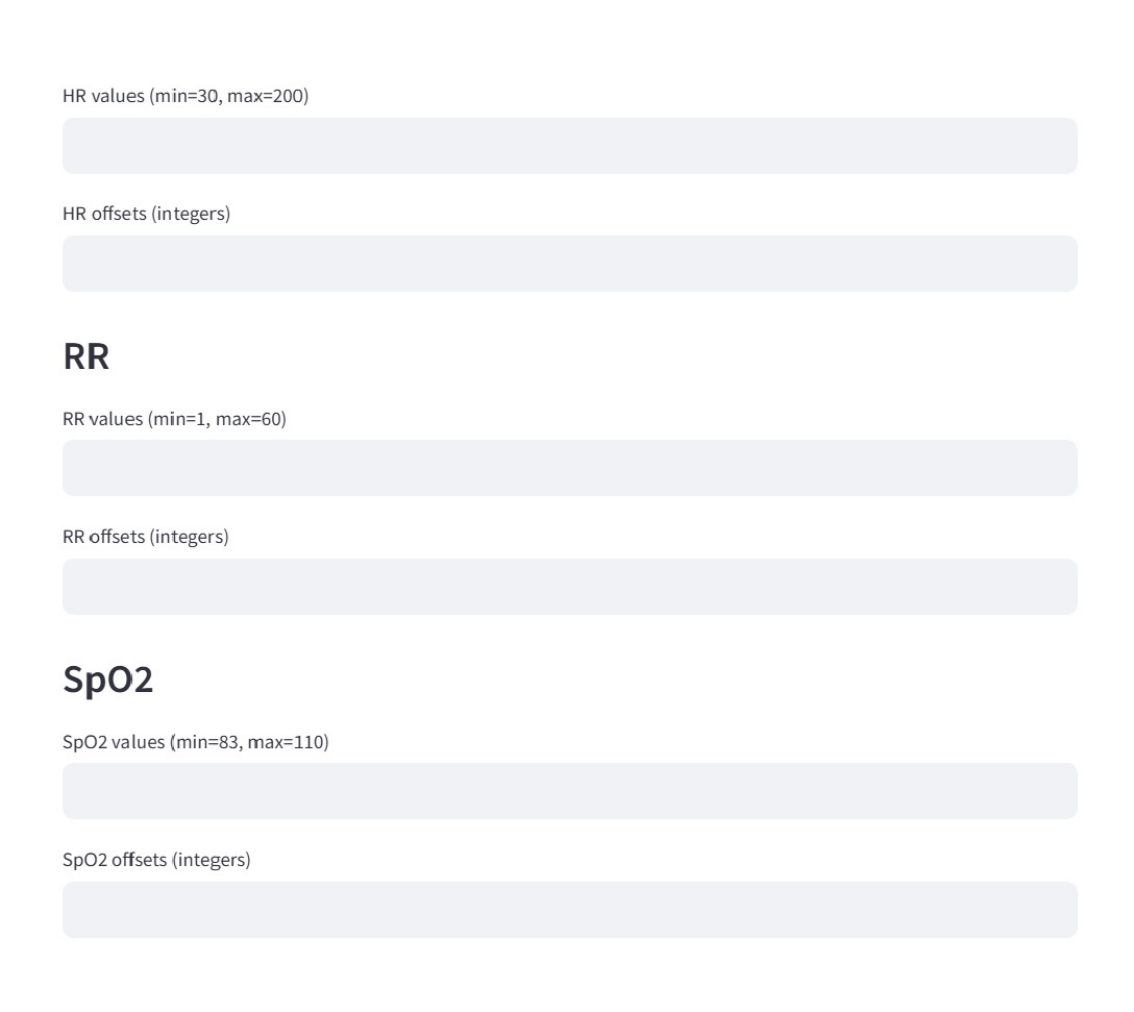}
    \caption{Vital signs tab. Users enter comma-separated values and time offsets for each signal, reflecting irregular time-series format of real ICU data.}
    \label{fig:vital_sign_tab}
\end{figure*}

\begin{itemize}
    \item \textbf{Vital Signs:} High-frequency signals like HR, RR, SpO2, blood pressures. Users enter comma-separated numeric values with corresponding timestamp offsets (see Figure~\ref{fig:vital_sign_tab}).
    \item \textbf{Medications:} Discrete entries for medications and infusions, including categorical fields (e.g., route, frequency) selected via drop-downs, and numeric dosage values entered via sliders.
    \item \textbf{Lab Results:} Similar to medications, lab name and result are entered, along with timestamp offset.
    \item \textbf{Scores:} Includes GCS and sedation scores. Each record consists of a categorical score and offset.
    \item \textbf{IO and IO Total:} IO tab supports both discrete and total values. Total intake/output values are entered similarly to vital signs, while IO cell types (e.g., dialysis) are entered individually.
    \item \textbf{Past History and Treatment:} Categorical variables like comorbidities and treatments are entered one by one with drop-downs.
    \item \textbf{Diagnosis:} Supports a broad set of diagnoses and priorities, and also separate ``admission diagnosis'' (addx).
    \item \textbf{Demographics and Hospital Info:} Includes static features like weight, height, gender, age, and hospital and unit metadata.
\end{itemize}

Several tabs include \textbf{multiple data sources} grouped logically (e.g., the ``Vital Signs'' tab includes HR, RR, SpO2, BP measurements). This design improves usability and visualization.

\subsection*{Notes on Limitations and Implementation}
\begin{itemize}
    \item  \textbf{Min and Max Limits:} All numeric sliders (e.g., for dosage or lab values) are bounded using thresholds based on the mean $\mu$ and the standard deviation $\sigma$ calculated from training data. These are \textit{not clinically validated limits} and are used purely for demonstration.
    \item \textbf{Manual Data Entry:} In a real-world scenario, clinicians and nurses would not manually input most of the fields. Many values such as vitals and labs are collected automatically through hospital systems and monitoring equipment. Only specific fields (e.g., diagnosis or treatments) may require human input.
    \item \textbf{Efficiency:} The current implementation is not optimized for performance and is intended for proof-of-concept only. Future work may focus on inference time improvements and front-end optimizations.
    \item \textbf{Deployment:} The model is pretrained offline and deployed for inference only. All processing happens server-side using CPU.
\end{itemize}

\clearpage

\clearpage

\clearpage

\end{document}